\documentclass[lettersize,journal]{IEEEtran}
\usepackage{amsmath,amsfonts}
\usepackage{algorithmic}
\usepackage{array}
\usepackage[caption=false,font=normalsize,labelfont=sf,textfont=sf]{subfig}
\usepackage{textcomp}
\usepackage{stfloats}
\usepackage{url}
\usepackage{verbatim}
\usepackage{graphicx}
\def\BibTeX{{\rm B\kern-.05em{\sc i\kern-.025em b}\kern-.08em
    T\kern-.1667em\lower.7ex\hbox{E}\kern-.125emX}}
\usepackage{balance}
\pdfoutput=1
\usepackage{bbding}
\usepackage{multirow}
\usepackage{tabu}
\usepackage{bm}
\usepackage{algorithm}

\usepackage{cite}
\usepackage{color}
\usepackage{booktabs}
\usepackage{verbatim}
\definecolor{darkgreen}{rgb}{0,0.5,0.2}
\definecolor{darkblue}{rgb}{0,0.3,0.6}

\usepackage{booktabs}

\usepackage{amsmath}
\usepackage{amssymb}
\usepackage{xcolor}
\usepackage{adjustbox}
\usepackage{makecell}
\usepackage{textcomp}
\usepackage{listings}
\usepackage{pifont}
\usepackage{colortbl}
\DeclareMathOperator*{\argmax}{arg\,max}

\definecolor{lightgray}{rgb}{0.835, 0.835, 0.835}
\definecolor{lightergray}{rgb}{0.935, 0.935, 0.935}
\definecolor{codeblue}{rgb}{0.25,0.5,0.5}
\definecolor{keyword}{rgb}{0.8, 0.25, 0.5}
\definecolor{darkgreen}{RGB}{0,110,0}
\definecolor{darkred}{RGB}{170,0,0}
\newcommand{\note}[1]{\textcolor{black}{#1}}

\def\greencheckmark{\textcolor{darkgreen}{\checkmark}}
\def\redxmark{\textcolor{darkred}{\text{\ding{55}}}}  %

\newcommand{\listingsttfamily}{\fontfamily{pcr}\small}
\begin{document}
\title{ZeroPose: CAD-Prompted Zero-shot Object 6D Pose Estimation in Cluttered Scenes}
\author{
  Jianqiu Chen,
  Zikun Zhou,
  Mingshan Sun,
  Rui Zhao,\\
  Liwei Wu,
  Tianpeng Bao,
    Zhenyu He~$^\dagger$~\IEEEmembership{Senior Member, IEEE}
\thanks{
    Jianqiu Chen and Zhenyu He (Corresponding author $\dagger$) are with the School of Computer Science and Technology, Harbin Institute of Technology, Shenzhen, China (e-mail: zhenyuhe@hit.edu.cn). Zhenyu He is also with Pengcheng Laboratory, Shenzhen, China.  Zikun Zhou is with Pengcheng Laboratory, Shenzhen, China. Mingshan Sun, Tianpeng Bao,  Rui Zhao, and Liwei Wu are with SenseTime Research.
    }}

\markboth{Journal of \LaTeX\ Class Files,~Vol.~18, No.~9, September~2020}%
{How to Use the IEEEtran \LaTeX \ Templates}

\maketitle
\begin{abstract}
    Many robotics and industry applications have a high demand for the capability to estimate the 6D pose of novel objects from the cluttered scene.
    However, existing classic pose estimation methods are object-specific, which can only handle the specific objects seen during training. 
    When applied to a novel object, these methods necessitate a cumbersome onboarding process, which involves extensive dataset preparation and model retraining. 
    The extensive duration and resource consumption of onboarding limit their practicality in real-world applications
    In this paper, we introduce ZeroPose, a novel zero-shot framework that performs pose estimation following a Discovery-Orientation-Registration (DOR) inference pipeline. This framework generalizes to novel objects without requiring model retraining.
    Given the CAD model of a novel object, ZeroPose enables in seconds onboarding time to extract visual and geometric embeddings from the CAD model as a prompt. 
    With the prompting of the above embeddings, DOR can discover all related instances and estimate their 6D poses without additional human interaction or presupposing scene conditions.
    Compared with existing zero-shot methods solved by the render-and-compare paradigm, the DOR pipeline formulates the object pose estimation into a feature-matching problem, which avoids time-consuming online rendering and improves efficiency.
    Experimental results on the seven datasets show that ZeroPose as a zero-shot method achieves comparable performance with object-specific training methods and outperforms the state-of-the-art zero-shot method with 50x inference speed improvement.

\end{abstract}
\section{Introduction}
\label{Introduction}
\begin{figure}
    \centering
    \includegraphics[scale=0.68]{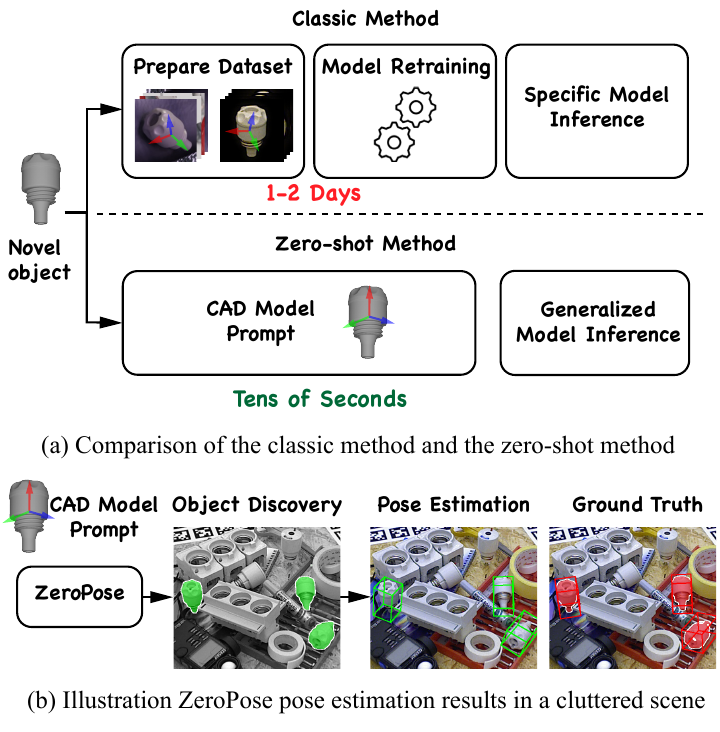}
    \caption{(a) The classic pose estimation method applied on a novel object needs a cumbersome onboarding process for preparing a training dataset and retraining an object-specific model. The zero-shot pose estimation method adopts a pre-trained generalized model without model retraining for specific objects, reducing the onboarding time from days to tens of seconds. (b) With the CAD model prompting, the proposed ZeroPose solves both object discovery and pose estimation in a zero-shot manner. }
    \label{fig:mainfig}
\end{figure}

\IEEEPARstart{P}{ose} estimation is a fundamental task for bin-picking or robot-grasping in the manufacturing field. It not only involves detecting objects in 3D space but also accurately estimating the six degrees of freedom pose transformation, including the relative orientation and position, \emph{w.r.t.} the defined CAD model.
Based on the estimated pose transformation, we can establish the point-level correspondences between the given CAD model and real-world observations. Such correspondences enable robots to interact with their environment in a more informed and safe manner, especially for applications that require high precision, including collision detection, and grasp point detection in the robotic arm control.

Classic solutions~\cite{densefusion, posecnn, hodavn2020bop, chen2023geo6d, Wang_2021_self6dpp, feng2023nvr, Cao_2023_DGECN_tcsvt} for pose estimation opt to learn a specific model for each object of interest using corresponding data to predict the pose precisely and robustly. However, when encountering an unseen object, they have to collect meticulously annotated data and then train a specific model for the unseen object, as shown in Figure~\ref{fig:mainfig}(a). Collecting thousands of training samples and retraining a model is time-consuming (usually takes 1 or 2 days) and necessitates the efforts of professional engineers. The high onboarding costs restrict the broad application of 6D object pose estimation in robotics and industrial fields.	

To reduce the onboarding costs, several studies~\cite{okorn2021zephyr, park2020latentfusion, labbe2022megapose, xie2021unseen, cai2022ove6d, shugurov2022cvpr:osop} pay attention to zero-shot object 6D pose estimation, which aims to empower the model with the ability to handle the object unseen training dataset, instead of retraining a new model for the novel object. 
Nevertheless, the zero-shot setting places rigorous demands on the generalization of the model. These algorithms rely on either additional human interactions or presupposing scene conditions, which limit the practicality in real-world applications.

Several algorithms~\cite{labbe2022megapose, okorn2021zephyr, xie2021unseen, cai2022ove6d} rely on human interactions to manually localize all instances of objects from the scene and estimate their 6D poses by the model, showing limited practicality. In addition, these pose estimation models~\cite{labbe2022megapose, okorn2021zephyr, xie2021unseen, cai2022ove6d} have an efficiency and effective issue from the render-and-compare paradigm. The computation complexity in this paradigm scales linearly with the number of candidate templates, leading to high computation costs and low inference speed. Besides, this paradigm is sensitive to illumination inconsistency, damaging the generalization ability of illumination conditions.
The recent OSOP method~\cite{shugurov2022cvpr:osop} proposes a pipeline enabling automatic object discovery under specific scene conditions. This pipeline employs a zero-shot semantic segmentation module, capable of performing object discovery under the assumption of only one instance for each target object within the scene.
However, object discovery by semantic segmentation is insufficient to identify the multiple instances of an object facing the challenge in cluttered scenes. 

\begin{figure*}[t]
    \centering
    \includegraphics[width=1.0\textwidth]{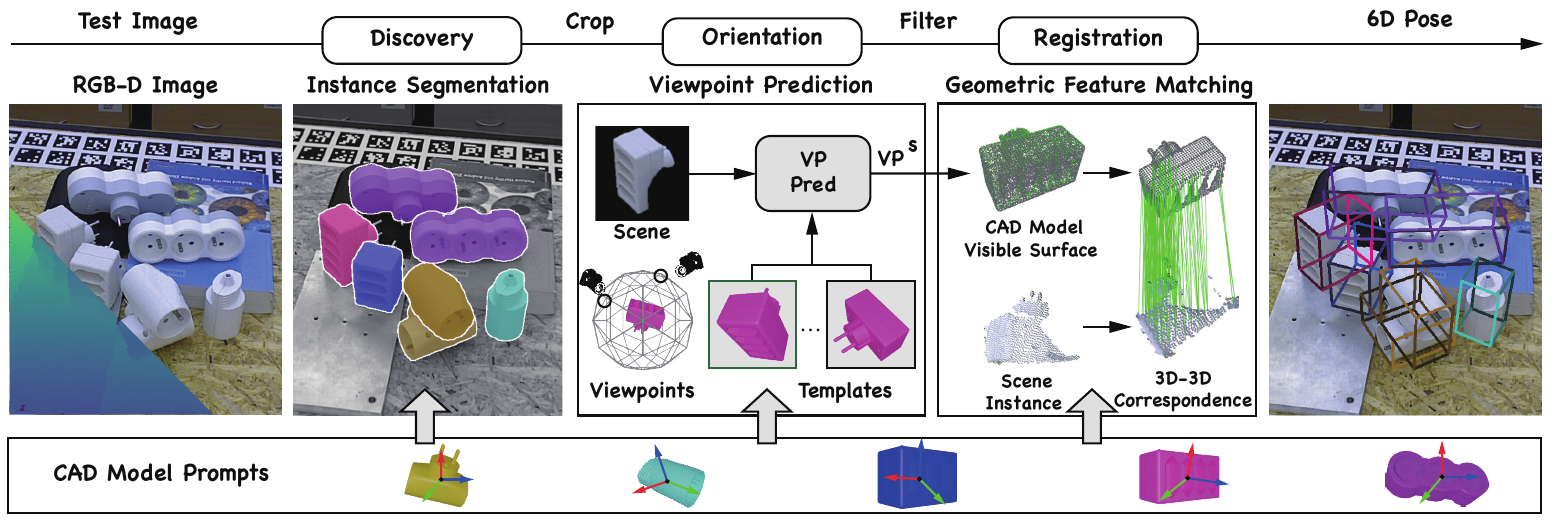}
    \caption{A high-level overview of ZeroPose. With the prompting from CAD models, ZeroPose enables both object discovery and pose estimation in a zero-shot manner following a Discovery-Orientation-Registration (DOR) inference pipeline.}
    \label{fig:framework}
\end{figure*}

In this paper, we propose a universal framework called ZeroPose, which solves both object discovery and pose estimation in a zero-shot manner without additional human interaction or presupposing scene conditions. It performs pose estimation following a Discovery-Orientation-Registration (DOR) inference pipeline on the RGB-D image, and these three inference steps are all achieved via feature matching in which the CAD model is used as a reference. Therefore, our ZeroPose can generalize to the unseen object given its CAD model. Given the CAD model of an unseen object, ZeroPose begins with extracting its visual and geometric embeddings, which are further used as prompts. The process of the CAD model is also known as onboarding. ZeroPose then predicts the 6D pose for each instance of the object following the DOR pipeline with the prompting of the above embeddings. Figure~\ref{fig:framework} showcases the DOR inference process within a cluttered scene, delineating each step along with its corresponding results.

The first step of DOR is to discover all instances of the target object from the cluttered scene. We resort to the Segment Anything Model~\cite{sam,fastsam} for discovering all foreground instances and then conducting feature matching between these instances and the CAD model prompt, segmenting the instances belonging to every target object. After lifting the instance segmentation results into 3D, we conduct point matching between the instance point cloud in the scene and the CAD model to solve the zero-shot pose estimation task. However, it is a challenging task since the instance point cloud is incomplete due to self-occlusion. Particularly, the points in the CAD model corresponding to the occluded region increase the risk of mismatching. To address this issue, we divide the matching into two steps: Orientation and Registration. Herein the Orientation step is to estimate the camera observation viewpoint to find the points of the CAD model corresponding to the visible points of the instance and filter out the remaining points of the CAD model. Based on the predicted orientation, the Registration step performs point matching between the instance point clouds and the filtered CAD model to estimate the pose transformation.

We evaluate the proposed ZeroPose on the seven datasets of the BOP benchmark~\cite{hodavn2020bop} where there are over a hundred objects with a variety of shapes and textures and twenty thousand images under different scenes. Experimental results show that ZeroPose as a zero-shot method achieves comparable performance with object-specific training methods and shows 50x running speed improvement compared with the state-of-the-art zero-shot method.

% 实验结果
In summary, our paper makes the following contributions:
\begin{itemize}
\item \note{We propose a universal zero-shot pose estimation framework, ZeroPose, which employs a three-step Discovery-Orientation-Registration pipeline. This framework can generalize to novel objects without the need for model retraining or prior assumptions about scene conditions.}
\item \note{we build a CAD model prompted zero-shot instance segmentation module based on SAM, which first leverages SAM to generate all possible proposals and then associates the potential proposals with the CAD models.}
\item \note{We introduce a lightweight step-wise pose estimation paradigm that simplifies the challenge of 6D object pose estimation into camera viewpoint prediction and point cloud registration tasks.}
\end{itemize}
\section{Related Work}

\subsection{Zero-Shot Instance Segmentation}
The task of instance segmentation is designed to discover all instances of target objects within a cluttered scene. In the field of pose estimation, instance segmentation also serves as a crucial preliminary step. Many pose estimation methods~\cite{su2022zebrapose, haugaard2022surfemb, labbe2022megapose} take instance segmentation as input to estimate the 6D poses of each detected instance.

Existing zero-shot instance segmentation methods~\cite{zheng2021zero, groundedsam} are typically based on the text prompt as the reference of the target objects. For the dataset with the common target objects such as COCO~\cite{coco}, text-prompt is effective to be understood by the language model and aligned with the visual feature from the scene for segmentation. However, for the pose estimation dataset, objects with highly customizable attributes and minor inter-object variations. These challenges make it difficult to distinguish objects through textual descriptions.

Besides the text-prompt zero-shot instance segmentation, there are some related works for object discovery. Previous research~\cite{xie2021unseen,shugurov2022cvpr:osop, xiang2021learning, ornek2023supergb, sam} has primarily concentrated on zero-shot semantic segmentation and zero-shot category-agnostic instance segmentation. 
Some methods~\cite{xie2021unseen, ornek2023supergb} leverage the RGB and depth image for unseen object instance segmentation in a semantic category-agnostic manner.  
Recently, the promptable zero-shot instance segmentation methods~\cite{sam, fastsam} are proposed, enabling segmenting instances under various types of prompts, \textit{e.g.}, points, boxes, and anything in the foreground. However, these methods are primarily designed for foreground instance segmentation, unable to associate the predicted instances to the candidate target objects.
In summary, the existing text or point prompts zero-shot instance segmentation methods are insufficient to distinguish the target objects in the pose estimation task. 
% There is a lack of an effective zero-shot method for object discovery in the pose estimation task.

\subsection{Unseen Object 6D Pose Estimation}
\textbf{Category-level pose estimation} is proposed to alleviate the expensive dataset preparation and training cost in recent methods ~\cite{lin2022icra:centerpose, wang2019normalized, liu_2022_cl6d }. The target objects in these methods are divided into categories and the model is trained for generalization on these categories. 
During inference, the model can be generalized to the novel object belonging to the categories seen in the training, eliminating the need for additional training.
However, the applicability of this category-level pose estimation approach is limited when encountering objects of categories not present in the training dataset or when there are large intra-category variations in shape and appearance.

\textbf{CAD-model-free pose estimation} methods ~\cite{sun2022cvpr:onepose, he2022oneposeplusplus, he2022cvpr:fs6d} 
leverage \textbf{one-shot} or \textbf{few-shot} pose-annotated images of the object as a reference to estimate the object pose in query images as the ``virtual anchors" of Augmented Reality (AR) effects. 
Since target objects in AR applications are arbitrary household objects from daily lives, the CAD model is unknown and unnecessary for establishing the point-level correspondence like demands in robotic applications. 
Therefore, annotating the object poses by the user in a few scene images is an adorable and effective way to perform pose estimation in AR applications. 
\note{Additionally, the pose of specific targets, such as humans, can be estimated using prior information like human skeletons~\cite{hpe_tmm}. Recent works propose a multi-stage pipeline~\cite{hpe_mm18} and decentralized pose representation~\cite{hpe_mm23} to successfully estimate multi-person poses in challenging crowded scenes.}

However, since the pose-annotated image prompt does not have a high-fidelity shape definition, these methods can not establish the important point-level correspondences from the pose-annotated image prompt for grab points detection or collision detection tasks.  
Moreover, the user annotation prompt reduces automation and practicality.

\begin{figure*}[t]
    \centering
    \includegraphics[width=1.0\textwidth]{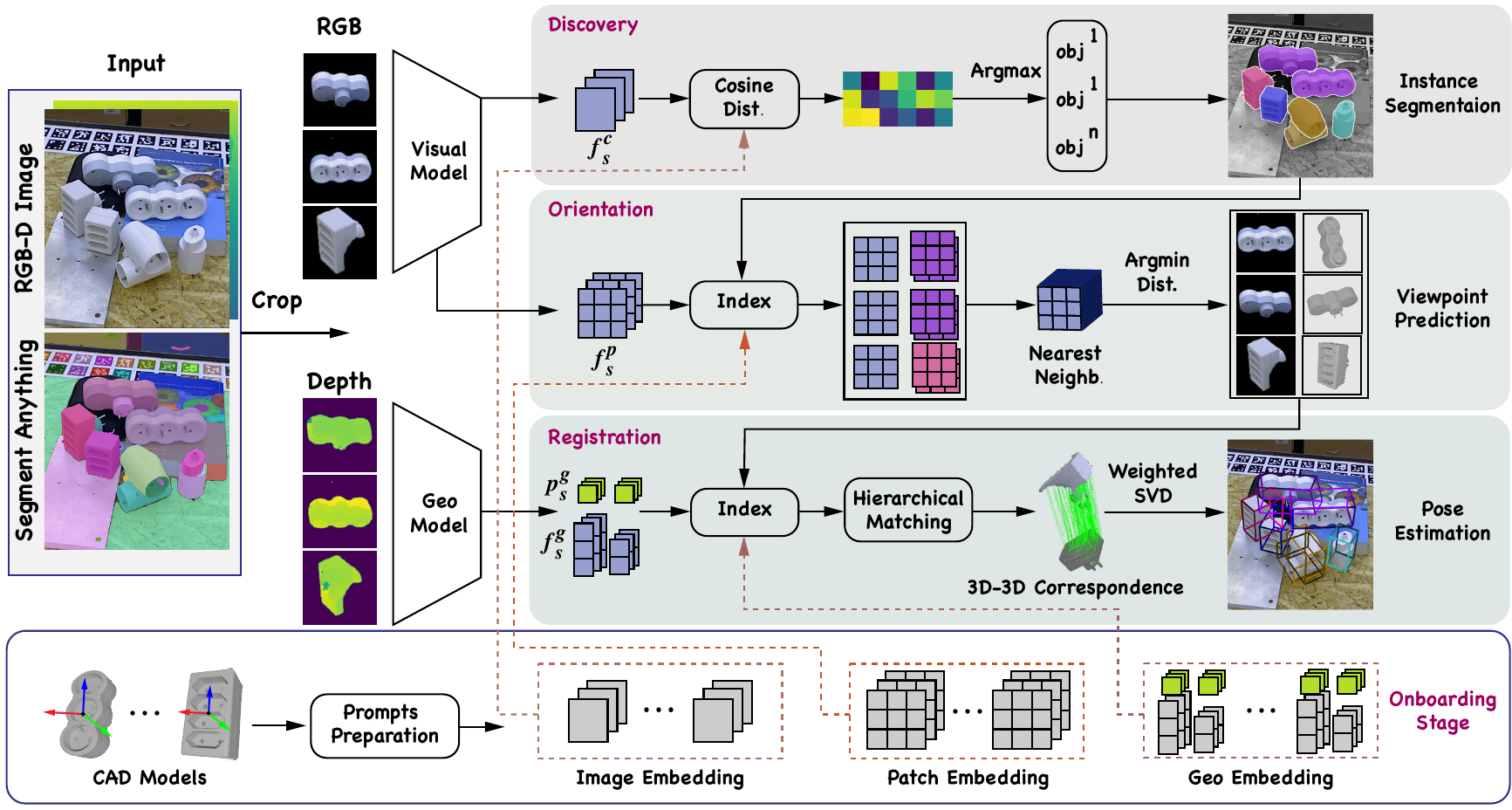}

    \caption{\note{An illustration of ZeroPose. 
    ZeroPose begins with extracting embeddings from the CAD model as prompt. With the prompting, the Discovery-Orientation-Registration (DOR) inference pipeline achieves pose estimation at three inference steps.
    The Discovery step aims for object discovery from the cluttered scene. It calculates the image embedding cosine similarity between foreground instances from scene segment anything results and CAD model prompts and associates them based on the cosine similarity score.
    The Orientation step is to estimate the camera observation viewpoint to find the points of the CAD model corresponding to the visible instance points. It is based on the discovery results to index related CAD model patch embedding and estimates the camera observation viewpoint by the nearest neighbor patch embedding feature distance.
    The Registration step solves the pose transformation by geometric embedding matching between the instance point clouds and the filtered CAD model point clouds. }}
    \label{fig:pipeline}
\end{figure*}
\textbf{CAD-model-based pose estimation} is introduced in recent methods ~\cite{park2020latentfusion, okorn2021zephyr, labbe2022megapose, cai2022ove6d, Zhou_2021_ss6d_tcsvt, Cao_2023_DGECN_tcsvt, sun2023uni6dv2, hff6d}. 
Since the CAD model is known as the pose and shape definition of the target object and does not require any annotated samples from the real scene, these methods are usually seen as \textbf{zero-shot} learning methods. The existing zero-shot methods employ a rendering-and-compare pipeline for the generalization of different novel objects. Given a CAD model of a novel object, the pipeline online renders numerous pose hypotheses as templates for each instance of the scene and selects the optimal pose with the highest appearance feature similarity.
However, the computational consumption and inference runtime increase linearly with the number of potential templates presenting an efficiency challenge. Moreover, the instances are required to be cropped from the scene by human interaction, which limits the practicality and results in a partial zero-shot setting.

Recent method OSOP~\cite{shugurov2022cvpr:osop} achieves the zero-shot pipeline under a presupposing scene condition with one and only one instance of the target object in the scene. This pipeline employs a zero-shot semantic segmentation module, capable of performing object discovery under the assumption of only one instance for each target object within the scene. However, for the cluttered scene such as in the T-LESS dataset~\cite{tless}, object discovery by semantic segmentation is insufficient to discover multiple instances of an object.

Besides, we analyze a potential zero-shot pose estimation approach, point cloud registration~\cite{drost2010model, qin2023geotransformer, pcr_tcsvt, cspcr_tcsvt}. It is typically designed for scene-level registration to estimate the camera pose transformation between two scene images. 
However,  due to the capturing source gap and a wide range of object scales, existing registration methods underperform on real-world object pose estimation datasets~\cite{hodavn2020bop}, as evidenced in the literature~\cite{matchnorm}. The capturing source gap of the scene and CAD model (RGB-D camera and 3D scanning) leads to the ratio of matchable regions less than 50\%. There are only a few points as reliable inliers for matching, and the rest of the points in the CAD model are outliers leading to a mismatch and failure pose transformation. 
Moreover, the wide range of object scales introduces another challenge to generalization capability. Although the object pose estimation method MatchNorm~\cite{matchnorm} introduces an additional normalization layer for generalizing to different object scales, the learnable layer in MatchNorm requires training for specific objects limiting its practicability.

The proposed ZeroPose introduces a novel DOR inference pipeline addressing the 6D object pose estimation through feature matching, which avoids the need for time-consuming online rendering and improves computational efficiency. Following the release of our preprint, the significance and generalization potential of zero-shot pose estimation has garnered increased attention within the research community~\cite{nguyen2023cnos, sam6d, pomz}, underscoring the impact of ZeroPose.
\section{Method}
\label{sec:method}

\begin{figure*}[t]
    \centering
    \includegraphics[width=1.0\textwidth]{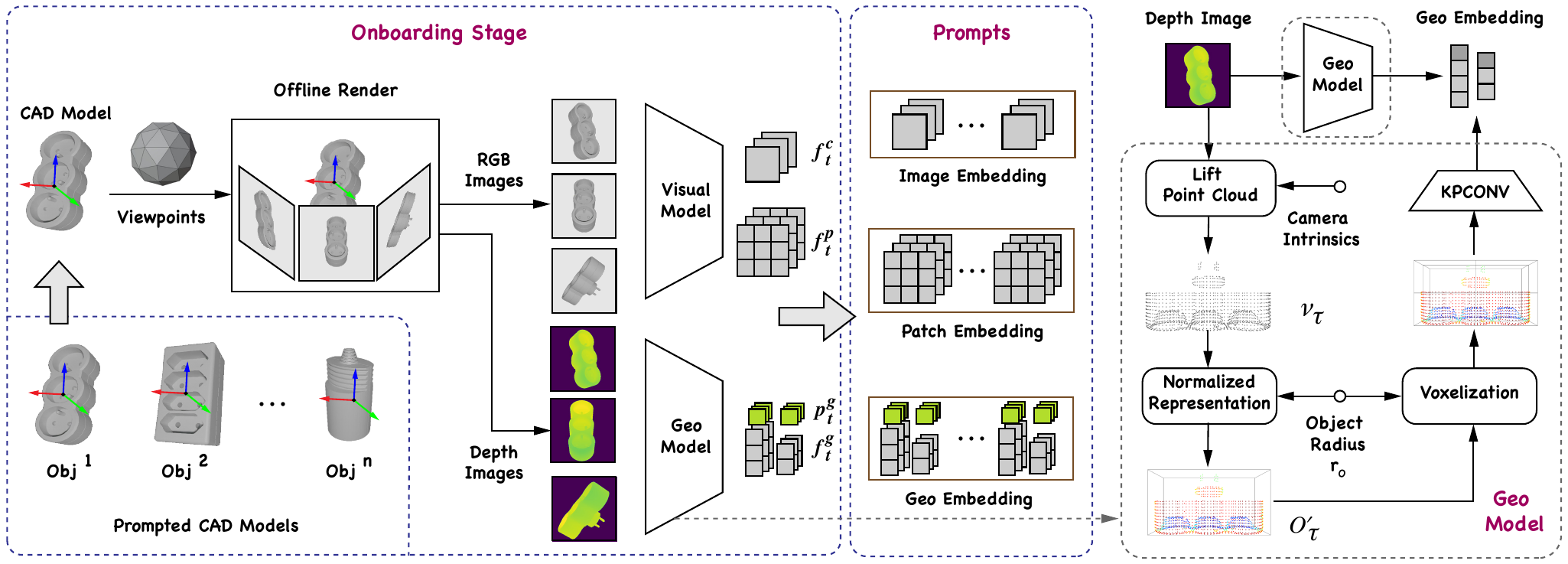}
    \caption{\note{Left: The onboarding stage aims to extract visual and geometric embeddings from the CAD model as the prompt of the target object, which is offline and only requires running once for each object. Right: Illustration for the Geo Model.}}
    \label{fig:onbard_geomodel}
\end{figure*}

\begin{figure}[t]
\centering
\centerline{{\includegraphics[width=0.49\textwidth]{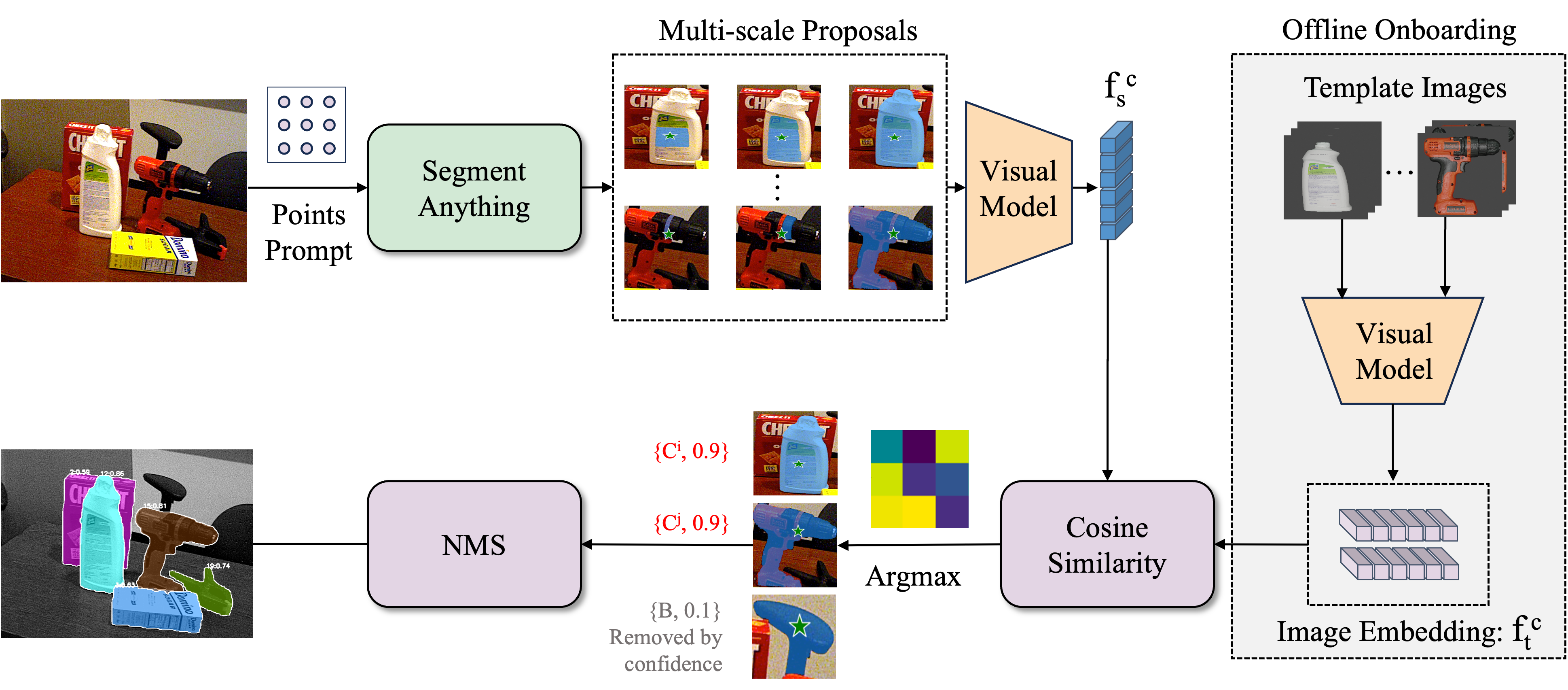}}}
\caption{\note{An illustration of the discovery step of ZeroPose.}}
\label{fig:discovery_detailed}
\end{figure}

\begin{figure}[t]
\centering
\centerline{{\includegraphics[width=0.49\textwidth]{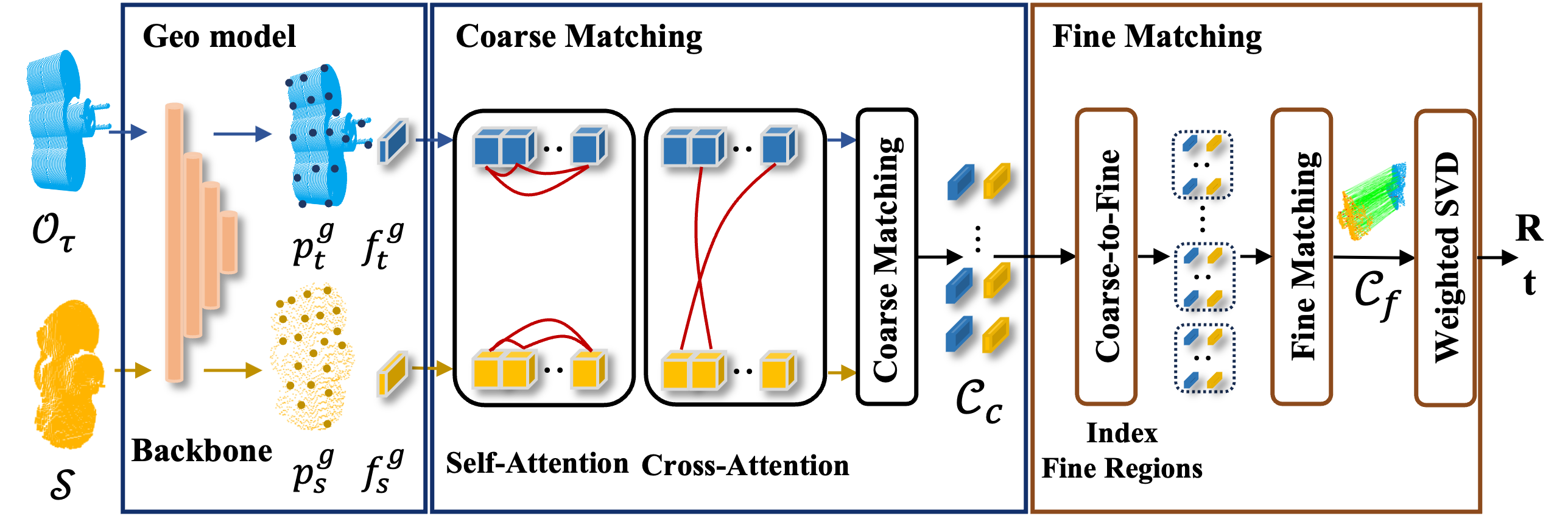}}}
\caption{\note{The architecture of hierarchical matching in the Registration step.}}
\label{fig:r2_vis_hm}
\end{figure}

Figure~\ref{fig:pipeline} illustrates the ZeroPose framework. The goal of ZeroPose is the discovery of all instances of target novel objects and estimate their 6D object poses relative to the CAD model. To achieve that, given the CAD models, ZeroPose begins with preparing visual and geometric embeddings from the CAD model as prompts, at the onboarding stage (Section~\ref{sec:method_onbard}). At inference, ZeroPose performs pose estimation following a Discovery-Orientation-Registration (DOR) inference pipeline with the prompting of the above embeddings. 
The discovery step (Section~\ref{sec:method_seg}) is to segment and crop all instances of objects from the cluttered scene and associate their related CAD models by image embedding feature matching. 
The orientation step (Section~\ref{sec:method_vp}) leverages the point embedding matching to estimate the camera observation viewpoint to find the points of the CAD model corresponding to the visible points of the scene instance.
The registration step (Section~\ref{sec:method_pose}) aims to estimate the pose transformation from the scene instance point clouds and filtered CAD model point clouds by geometric embedding feature matching. 

\subsection{Onboarding Stage}
\label{sec:method_onbard}
The CAD model serves as the reference for the orientation and position of the target object, encompassing diverse visual and geometric information. To leverage the information as a prompt, we introduce a strategy to extract both visual and geometric embeddings from the CAD model during the onboarding stage. As depicted in Figure~\ref{fig:onbard_geomodel}, we initially render the CAD model from various camera observation viewpoints, as outlined in Section~\ref{sec:method_vp}, resulting in $R$ template RGB-D images and extract embeddings from the template images.

For extracting visual embeddings, a visual foundation model DINOv2~\cite{dinov2} (denoted visual model in Figure~\ref{fig:onbard_geomodel}), is utilized to extract visual embeddings from the template images. The visual embeddings consist of the image embedding and the patch embedding. The image embedding is an image-level visual feature presented in $f_t^c$ with shape $(R, C)$, where $R$ is the number of rendered template images, and $C$ is the feature dimension for the visual foundation model. The patch embedding is the patch-level visual feature presented in $f_t^p$ with shape $(R, P_l, C)$, where $P_l = P_h * P_w$, $P_h$, $P_w$ are the height and width of the patch-level visual feature output of visual model~\cite{dinov2}. 
\note{For extracting geometric embeddings, we employ a pretrained geometric model (referred to as the Geo model) to extract point cloud and geometric features from the CAD model. To achieve that, template RGB-D images are lifted into a color point cloud with normalized representation, and voxelize the point cloud $p_t^g$ to extract hierarchical geometric features $f_t^g$. The geometric embedding consists of $\{p_{t}^g, f_{t}^g\}$, where $p_t^g = \{p_{t, c}^g, p_{t, f}^g\}$ and $f_t^g = \{f_{t, c}^g, f_{t, f}^g\}$. $c$ and $f$ denote the coarse-level and the fine-level.}

Both operations in the onboarding stage are conducted offline and performed only once for each novel object before the inference stage.

\subsection{Discovery}
\label{sec:method_seg}
\note{Object discovery, as the first step in the DOR inference pipeline, aims to discover all instances of objects from the scene. 
To achieve zero-shot object discovery, we resort to a Segment Anything Model (SAM)~\cite{fastsam} to generate all possible proposals and then associate the potential proposals with the CAD models via feature matching.}

\note{
For proposal generation, we leverage the SAM with a uniform point set as a prompt to generate proposal instance masks. 
Since the proposal instance masks are object-agnostic, the proposal instances are needed to associate with the prompted CAD model, and irrelevant instances need to be filtered out.
To this end, we propose an object association approach, demonstrated in Figure~\ref{fig:discovery_detailed}.
Specifically, we crop these instances from the image and resize them into a fixed image size for extracting their image embedding through the visual model. Image embedding is the image-level visual feature $f_s^c$, with dimensions $(M, C)$. $M$ represents the number of proposal instances. 
Then, we calculate the global visual feature similarity between the proposal instances and prepared template images. 
The proposal instance is scored by this highest similarity and assigned with the ID of the CAD model corresponding to the highest feature similarity template images.
The association processing is formulated as follows:
\begin{equation}
O_{id} = \argmax_{n=1,...,N} \max_{\mathcal{T} = 1, ..., R} \left( \frac{f_{s}^c \cdot f_{t, n, \mathcal{T}}^c}{\|f_{s}\| \|f_{t, n, \mathcal{T}}^c\|} \right),
\end{equation}
where $O_{id}$ is the related CAD model ID of scene instances. The instance mask combined with the predicted CAD model ID results in zero-shot instance segmentation.}

\note{To filter irrelevant instances, we remove the proposal instances by score threshold. Besides, for some over/under-segmentation results, we leverage the Non-Maximum Suppression (NMS) to filter out. That is because the scoring mechanism will assign quite a low score for the over/under-segmentation result and assign a higher score for the accurate-segmentation result. Therefore, we filter out the suboptimal segmentation masks by NMS.}

Additionally, we introduce a text-prompt zero-shot instance segmentation within pose estimation datasets for comparison. There is a similar pipeline that instead of the image embedding into the text embedding. This pipeline leverages the multi-modal vision model~\cite{imagebind} instead of the visual model~\cite{dinov2} to extract the multi-modal feature of the scene image feature and the text feature of target objects for matching.
However, in the pose estimation dataset, many objects are manufactured often without text descriptions for embedding.
To solve this issue, we leverage the point cloud foundation modal PointLLM~\cite{xu2023pointllm} to generate a caption of the CAD model for extracting the text embedding. Details of the design and experiment results are shown in Section~\ref{sec:exp_seg}.

\subsection{Orientation}
\label{sec:method_vp}
The Orientation step aims to estimate the camera observation viewpoint to find the points of the CAD model corresponding to visible points of the predicted instance in the scene and filter out the remaining points of the CAD model. 
Since the instance in the scene is a partial observation from a camera viewpoint, it is incomplete due to self-occlusion. Directly estimating the 3D correspondence between the occluded scene instance and the CAD model increases the risk of mismatching.
To solve this, the orientation step predicts the camera observation viewpoint by patch embedding feature matching with a few discrete templates. Then, based on the pin-hole imaging principle, we index the points of the CAD model corresponding to the visible instance points and filter out the remaining points of the CAD model. 

\note{Compared with the 6 degrees of freedom (DoF) camera perspective rendering for pose estimation, the predefined camera viewpoints are tailored for filtering invisible points of the CAD model.
To reduce the number of rendering templates and improve efficiency, we only vary the visible region-related 2 DoF optical axis direction in camera perspective for template rendering and fix the rest 3 DoF translation and 1 DoF in-plane rotation.
Specifically, by revisiting the camera observation viewpoint, we find that the 3 DoF orientation of camera viewpoint $\mathbf{R}_{\mathcal{O}}$ can be decomposed into a 2 DoF optical axis direction $\mathbf{R}_{\gamma}$ of the image plane and a one-degree-of-freedom in-plane rotation $\mathbf{R}_{\theta}$ among this direction.} This decomposition is formalized by the equation:
\begin{equation}
    \mathbf{R}_{\mathcal{O}} = \mathbf{R}_{\gamma} \mathbf{R}_{\theta},
\end{equation}
where 
\begin{equation}
    \mathbf{R}_{\gamma} = \begin{bmatrix}
\cos \varphi \cos \psi & -\sin \varphi & \cos \varphi \sin \psi\\
\sin \varphi \cos \psi & \cos \varphi & \sin \varphi \sin \psi \\
-\sin \psi & 0 & \cos \psi \\
\end{bmatrix},
\end{equation}
\begin{equation}
    \mathbf{R}_{\theta} = \begin{bmatrix}
        \cos \beta  & -\sin \beta & 0\\
        \sin \beta & \cos \beta & 0\\
        0 & 0 & 1 
       \end{bmatrix},
       \label{eqn:r_ip}
\end{equation}
% The $\mathbf{R}_{\gamma}$ is viewpoint direction with two degrees of freedom represented in Euler angles ($\varphi$, $\psi$), and $\mathbf{R}_{\theta}$ is with one in-plane rotation degree of freedom for rotation angle $\beta$.
($\varphi$, $\psi$, $\beta$) are Euler angles.
\note{Since in-plane rotation does not affect the visible region, the $\mathbf{R}_{\mathcal{O}}$ prediction task can be simplified into a two-degree-of-freedom viewpoint direction $\mathbf{R}_{\gamma}$ prediction task.} The candidate camera viewpoints can be uniformly sampled from a sphere surface~\cite{nguyen2022template} and adopt a few discrete templates to approximate its distribution because of the narrowed sampling space.

\note{
To predict the closest viewpoint from templates associated with the detected object, we introduce patch-level feature matching in the viewpoint prediction step.
Since closed viewpoint template images only have minor and local appearance variations, the image embedding for global semantic information in the discovery step is insufficient to distinguish these similar viewpoint template images of an object. 
To solve that, we extract patch-level features from visual model~\cite{dinov2} as patch embedding to calculate candidate viewpoints by patch embedding matching between the instance in the scene and template images.
Specifically, given viewpoint template images and the scene instance predicted from the discovery step, we adopt a long-side resize to scale them into a fixed image size and hold the optical axis in the center. Given the cropped images, we extract patch-level features from visual model~\cite{dinov2} as patch embedding, which is rotate-invariant for in-plane rotation $\mathbf{R}_{\theta}$. 
Specifically, each patch region in the segmented proposal instance searches for the nearest patch in the reference template image and uses the mean of local maximum feature similarity as the score for selecting viewpoints as follows:
\begin{equation}
    % \mathbf{S_p^{\mathcal{T}}}= \frac{1}{N_s} \sum_{j=1}^{N_s} \min_{i=1,...,N_t^{\mathcal{T}}} ||f_{s,j}^p - f_{t, i}^p ||_2.
    \mathbf{S_p^{\mathcal{T}}} = \frac{1}{N_s} \sum_{j=1}^{N_s} \max_{i=1,\dots,N_t^{\mathcal{T}}} \frac{f_{s,j}^p \cdot f_{t,i}^p}{\|f_{s,j}^p\|_2 \|f_{t,i}^p\|_2}.
\end{equation}}
The template viewpoints are sorted by the local patch feature similarity $\mathbf{S_p^{\mathcal{T}}}$ and we select the top $k$ as candidates. 
\note{Besides, we also leverage the patch-level feature similarity to revise the predicted instance score and use the similarity to estimate the visibility of segmented instances.
Then, we update the score of each instance with local similarity and visibility via weighted summation. }
The visible region of the CAD model under these viewpoints can also be calculated from the depth image, as introduced in the onboarding stage Section~\ref{sec:method_onbard}.

\subsection{Registration}
\label{sec:method_pose}
The registration step aims to estimate rigid pose transformation $\mathbf{R} \in SO(3),\mathbf{\mathbf{t} \in \mathbb{R}^3}$ from the scene instance point clouds and filtered CAD model point clouds. %  by geometric embedding feature matching
However, recent research studies~\cite{lin2023relposepp, zhao2024dvmnet} have observed that the image feature matching for establishing the point-to-point correspondence is unreliable in the scenario of object pose estimation. The visual features are sensitive to the illumination inconsistency between the real scene and synthetic template images. To solve that, we introduce the geometric embedding feature matching. 
% As illustrated in Figure~\ref{fig:pipeline}, this step begins with a Geo model to extract hierarchical geometric features as geometric embedding from the scene and filtered CAD model point clouds. Then, a hierarchical matching module is proposed to estimate the point 3D-3D correspondence by hierarchical geometric embedding matching and formulate the pose estimation into a least-squares problem.
\note{As illustrated in Figure~\ref{fig:r2_vis_hm}, the Geo model hierarchically samples the point clouds and extracts their geometric features, which are then combined into a geometric embedding.
A hierarchical matching module subsequently utilizes these point clouds and their associated geometric features to facilitate feature interaction, aiming to find their 3D-3D correspondences. 
 Based on the predicted correspondence, we adopt the 3D coordinates of corresponding point pairs and the pose transformation formula to calculate the 6D pose $R$, $t$ by least-squares minimizing the transformed point cloud distances.}

\subsubsection{\textbf{Geo Model}} Geo model is to extract a stable geometric embedding that can be generalized for different scale and shape objects. There are three primary steps illustrated in Figure~\ref{fig:onbard_geomodel}.
The lift point cloud step is to lift the point cloud pairs from the scene and template depth images. For each instance in the scene, we crop and mask them from the image and lift their point cloud from the depth image given the camera intrinsic matrix, calculated as follows:
\begin{equation}
\begin{bmatrix}
x_i \\
y_i \\
z_i \\
\end{bmatrix}
=
\begin{bmatrix}
\frac{u_i - c_x}{f_x} \\
\frac{v_i - c_y}{f_y} \\
1 \\
\end{bmatrix}
\times d_i,
\end{equation}
where $c_x$, $c_y$, $f_x$, $f_y$ are from camera intrinsic parameters and $u_i$, $v_i$ is the coordinate in the 2D image and $d_i$ is the cooresponding depth value from the depth image.
Based on the segmentation result, we separately calculate each point cloud of instance $\mathcal{S}=\{\mathbf{s}_i \in \mathbb{R}^3 \mid i=1, \ldots, m\}$ from the depth image and combine their corresponding color from RGB image as texture feature.
For the filtered CAD model point cloud, we initially lift point cloud $\mathcal{V}_{\mathcal{T}}=\{\mathbf{v}_i \in \mathbb{R}^3 \mid i=1, \ldots, n\}$ from the template RGB-D image and undergo an inverse transformation using the template pose $T_{\mathcal{T}}$ to align to the base orientation
and position in the CAD model as follows:
\begin{equation}
\begin{bmatrix}
a_i \\
b_i \\
c_i \\
1 \\
\end{bmatrix}
=
T_{\mathcal{T}}^{-1}
\begin{bmatrix}
x_i^\mathcal{T} \\
y_i^\mathcal{T} \\
z_i^\mathcal{T} \\
1 \\
\end{bmatrix},
\end{equation}
where the ($x_i^\mathcal{T}, y_i^\mathcal{T}, z_i^\mathcal{T}$) represents the value of $p_i$ and ($a_i, b_i, c_i$) is the surface point $\mathbf{o}_i$ of the viewpoint-filtered CAD model $\mathcal{O}_{\mathcal{T}}=\{\mathbf{o}_i \in \mathbb{R}^3 \mid i=1, \ldots, n\}$.

To address scale variability in unseen objects, we propose a normalized representation in the following step.
Given the prompted CAD model, we calculate the radius $r_\mathcal{O}$ of the circumscribed sphere of the CAD model. 
For a point cloud pairs ($\mathcal{S}, \mathcal{O}_\mathcal{T}$), we scale both of them by the $\frac{1}{r_\mathcal{O}}$ into a normalized space, achieving ($\mathcal{S}', \mathcal{O}_\mathcal{T}'$). This normalized representation allows for the maintenance of a receptive field for geometric features, which can dynamically scale the point cloud into a shared normalized space.
Additionally, the normalized representation combines with centroid clustering enabling filtering of the noisy points from the sensor noise or inaccuracy segmentation. In the centroid clustering algorithm, the critical bandwidth parameters can be selected to 1 at any normalized representation instance point clouds, eliminating the need for manual hyperparameter tuning.

After normalization, the point clouds are inputted into the geometric backbone network to extract geometric features.
\note{We utilize KPCONV~\cite{thomas2019kpconv} as the backbone, a method that voxelizes the point cloud across various local receptive fields, thereby extracting hierarchical point clouds and geometric features as geometric embedding $\{p_{s}^g, f_{s}^g\}$, where $p_s^g = \{p_{s, c}^g, p_{s, f}^g\}$ and $f_s^g = \{f_{s, c}^g, f_{s, f}^g\}$. $c$ and $f$ denote the coarse-level and the fine-level.}
The Geo model is trained using correspondence losses together with the following geometric matching model.
\subsubsection{\textbf{Hierarchical Matching for Registration}} 
\note{In this step, the goal is to estimate 3D correspondences at geometric embeddings and estimate the final pose transformation parameters from the correspondences.
Given input geometric embeddings from scene instances and predicted template instances, the hierarchical matching module conducts coarse-to-fine feature interaction, and through the similarity of these features establishes the correspondence.
Then, we utilize the point cloud coordinates of correspondence features to fit a pose transformation that minimizes the distance between the transformed point clouds.}

\note{As shown in Figure~\ref{fig:r2_vis_hm}, given the input geometric embeddings from Geo model, we use the coarse-level point cloud coordinates and features from the scene instances $\{p_{s, c}^g, f_{s, c}^g\}$ and template instances  $\{p_{t, c}^g, f_{t, c}^g\}$ as input to perform the feature fusion and interaction by the attention modules and select the features with high similarity as coarse-level correspondence $\mathcal{C}_c$.}

\note{Then, we adopt the receptive field of coarse-level features in correspondence $\mathcal{C}_c$ to index related fine-level points and features. By refining correspondence in the fine-level feature matching, we estimate the dense point-to-point correspondence $\mathcal{C}_f$. 
Based on the dense correspondence $\mathcal{C}_f$ and point cloud coordinates, we can formulate the object pose estimation into a weighted least-squares problem.
We adopt the coordinates of the corresponding points to minimize the point cloud distance, as follows:
\begin{equation}
\label{equ:pose_transform}
\min _{\mathbf{R}, \mathbf{t}} \sum_{\left(\mathbf{o}_{xi}, \mathbf{s}_{yi}\right) \in \mathcal{C}_f}\left\|\mathbf{R} \cdot \mathbf{o}_{x_i}+\mathbf{t}-\mathbf{s}_{y_i}\right\|_2^2 ,
\end{equation}
where $\mathbf{s}_{yi}$ and $\mathbf{o}_{x_i}$ are matched corresponding points in the fine-level matching.
The closed-form solution of the least-squares problem is the pose parameters ${\mathbf{R}, \mathbf{t}}$ with minimal point cloud distance, which can be solved by the weighted singular value decomposition (weighted SVD)~\cite{svd} 
 algorithm.}

To enhance the robustness of viewpoint prediction, we introduce a multi-hypothesis registration approach that supports multiple candidate viewpoint-filtered point cloud input.
Given $k$ different viewpoint-filtered CAD model point clouds, we establish the 3D correspondences between the scene instance and them and result in $k$ candidate poses. Subsequently, these filtered CAD model point clouds are transformed into the scene according to their respective candidate poses. We calculate their Chamfer distances~\cite{chamfer} and select the final pose estimation result with minimal point cloud distances.

\section{Experiments}
\label{others}
\begin{table*}[tbp]
\begin{center}
\caption{\note{Instance segmentation results on the BOP challenge datasets. We report the mAP as a metric. The header ``Real" indicates training with real scene images. \dag denotes that the dataset is no real images provided for training. * denotes that the object name is not provided and we use the predicted caption from PointLLM~\cite{xu2023pointllm} as an alternative. }}
\label{tab:ZSIS result}
  \centering
  % \small %
 \setlength{\tabcolsep}{3.5pt}
\begin{adjustbox}{max width=\textwidth}
  \begin{tabular}{rlccccccccccc}
  \toprule
  & & \multicolumn{2}{c}{Settings} & \multicolumn{7}{c}{BOP7 Datasets} & & \\ 
 \cmidrule(lr){3-4} \cmidrule(lr){5-11}
   & Method & {\scriptsize \makecell[b]{Zero-shot}} & {\scriptsize \makecell[b]{Real}} & {\textsc{lm-o}} & {\textsc{t-less}} & \multicolumn{1}{c}{\textsc{tud-l}} & \multicolumn{1}{c}{\textsc{ic-bin}} & \multicolumn{1}{c}{\textsc{itodd}} & \multicolumn{1}{c}{\textsc{hb}} & \multicolumn{1}{c}{\textsc{ycb-v}} & Mean & Time \note{(s)}\\

    \midrule
     {\color{teal}\scriptsize 1} & Mask RCNN\cite{maskrcnn}  & $\redxmark$ & $\redxmark$  & 37.5 & 51.7 & 30.6 & 31.6 & 12.2 & 47.1 & 42.9 & 36.2 & 0.054\\
     {\color{teal}\scriptsize 2} & Mask RCNN\cite{maskrcnn}  & $\redxmark$ & $\greencheckmark$ & 37.5\dag & 54.4 & 48.9 & 31.6\dag & 12.2\dag & 47.1\dag & 52.0 & 40.5 & 0.055
 \\
     {\color{teal}\scriptsize 3} & ZebraPoseSAT\cite{su2022zebrapose} & $\redxmark$ & $\redxmark$  &50.6 & 62.9 & 51.4 & 37.9 & 36.1 & 64.4 & 62.6 & 52.3 & 0.080\\
     {\color{teal}\scriptsize 4} & ZebraPoseSAT\cite{su2022zebrapose} & $\redxmark$ & $\greencheckmark$ & 50.6\dag & 70.9 & 70.7 & 37.9\dag & 36.1\dag & 64.4\dag & 74.0 & 57.8 & 0.080\\
    \midrule
{\color{teal}\scriptsize 5} & Text-prompt (ImageBind) & $\greencheckmark$ & $\redxmark$ & 10.7 & 1.2* & 17.3* & 15.1 & 6.3 & 16.3* & 38.3 & 15.0 & 0.427 \\ 
     {\color{teal}\scriptsize 6} & CAD-prompt (ImageBind)  & $\greencheckmark$ & $\redxmark$ & 30.4 & 25.1 & 28.6 & 18.8 & 19.7 & 41.5 & 48.5 & 30.4 & 0.429\\ 
     {\color{teal}\scriptsize 7} & CAD-prompt (DINOv2)  & $\greencheckmark$ & $\redxmark$ & \textbf{37.7} & \textbf{34.7} & 46.0 & 20.6 & \textbf{20.2} & \textbf{47.8} & 57.4 & \textbf{37.8} & \textbf{0.220} \\
     {\color{teal}\scriptsize 8} & \note{CAD-prompt (DINOv2/Pyrender)}  & $\greencheckmark$ & $\redxmark$  & \note{34.4} & \note{31.3} & \note{\textbf{51.5}} & \note{\textbf{21.7}} & \note{14.6} & \note{47.6} & \note{\textbf{60.0}} & \note{37.1} & \note{\textbf{0.220}} \\
    \bottomrule      
\end{tabular}
\end{adjustbox}
\end{center}
\end{table*}
\begin{table*}[t]
  \centering
  % \small %
 \setlength{\tabcolsep}{3.5pt}

   \caption{\note{Pose estimation results on the BOP challenge datasets. We report the AR score on each of the 7 core datasets in the BOP challenge and the mean score across datasets. Zero-shot stands for the model in the Object Discovery or Pose Estimation step enabling generalization to the novel object without retraining. For each column, we denote the best over result in \textit{italics} and the best zero-shot pose estimation method for each setting block in \textbf{bold}. The unit of time column is seconds (s).}
  }
\begin{adjustbox}{max width=\textwidth}

  \begin{tabular}{rlcclccccccccccc}
 \toprule  
 & & \multicolumn{2}{c}{Object Discovery} & \multicolumn{2}{c}{Pose Estimation} & & \multicolumn{7}{c}{BOP7 Datasets} & & \\ 
 \cmidrule(lr){3-4} \cmidrule(lr){5-6} \cmidrule(lr){7-13}
  & Method & {\scriptsize \makecell[b]{Zero-shot}} & {\scriptsize \makecell[b]{Inst. level}} & {\scriptsize \makecell[b]{Zero-shot}} & {\scriptsize \makecell[b]{Refinement}} & \textsc{lm-o} & \textsc{t-less} & \textsc{tud-l} & \textsc{ic-bin} & \textsc{itodd} & \textsc{hb} & \textsc{ycb-v} & Mean & Time \note{(s)} \\

 \midrule %
{\color{teal}\scriptsize 1} & CosyPose~\cite{cosypose} & $\redxmark$ &$\greencheckmark$ & $\redxmark$ & $\greencheckmark$ &  63.3 & 64.0 & 68.5 & 58.3 & 21.6 & 65.6 & 57.4 & 57.0 & 0.5
\\ %
{\color{teal}\scriptsize 2} & CDPNv2~\cite{li2019cdpn} & $\redxmark$ &$\greencheckmark$ & $\redxmark$& $\greencheckmark$ &63.0 & 43.5 & 79.1 & 45.0 & 18.6 & 71.2 & 53.2 & 53.4 & 1.5
\\ %
{\color{teal}\scriptsize 3} & SurfEmb~\cite{haugaard2022surfemb} & $\redxmark$ &$\greencheckmark$ & $\redxmark$ & $\greencheckmark$ &  \textit{76.0} & \textit{82.8} & 85.4 & \textit{65.9} & \textit{53.8} & 86.6 & 79.9 & \textit{75.8} & 9.0
\\ %
{\color{teal}\scriptsize 4} & Coupled~\cite{lipson2022coupled} & $\redxmark$ &$\greencheckmark$ & $\redxmark$ & $\greencheckmark$ & 73.2 &	82.0 & 85.8 &	60.6	& 47.2	& \textit{87.3}	& 82.9 & 74.1 & -\\ %
 \midrule %

{\color{teal}\scriptsize 5} & MegaPose~\cite{labbe2022megapose} & $\redxmark$ &$\greencheckmark$ & $\greencheckmark$ & -- & 18.7 & 19.7 & 20.5 & 15.3 & 8.00 & 18.6 & 13.9 & 16.2 & 25.6 \\ %
{\color{teal}\scriptsize 6} & \note{OVE6D~\cite{cai2022ove6d}} & $\redxmark$ &$\greencheckmark$ & $\greencheckmark$ & \note{--} & \note{49.6} & \note{52.3} & \note{-} & \note{-} & \note{-} & \note{-} & \note{57.5} & \note{-} & \note{-} \\ %

{\color{teal}\scriptsize 7} & Ours & $\redxmark$ &$\greencheckmark$ & $\greencheckmark$ & -- & \textbf{58.3} & \textbf{55.9} & \textbf{86.9} & \textbf{53.2} & \textbf{33.8} & \textbf{69.3} & \textbf{70.1} & \textbf{61.1} & \textbf{1.8}
\\ %
 % \midrule %

{\color{teal}\scriptsize 8} & \note{OVE6D~\cite{cai2022ove6d}}& $\redxmark$ &$\greencheckmark$ & $\greencheckmark$ & $\greencheckmark$ & \note{62.7} & \note{54.6} & \note{-} & \note{-} & \note{-} & \note{-} & \note{58.7} & \note{-} & \note{-} \\ %

{\color{teal}\scriptsize 9} & \note{GCPose~\cite{gcpose}}& $\redxmark$ &$\greencheckmark$ & $\greencheckmark$ & $\greencheckmark$ & \note{65.2} & \note{\textbf{67.9}} & \note{92.6} & \note{-} & \note{-} & \note{-} & \note{75.2} & \note{-} & \note{-} \\ %

{\color{teal}\scriptsize 10} & MegaPose~\cite{labbe2022megapose} & $\redxmark$ &$\greencheckmark$ & $\greencheckmark$  & $\greencheckmark$  &58.3 & 54.3 & 71.2 & 37.1 & 40.4 & 75.7 & 63.3 & 57.2 & 93.3 \\ %

{\color{teal}\scriptsize 11} & Ours & $\redxmark$ &$\greencheckmark$ & $\greencheckmark$ & $\greencheckmark$ & \textbf{66.3} & 63.0 & \textit{\textbf{94.9}} & \textbf{52.0} & \textbf{44.2} & \textbf{82.0} & \textit{\textbf{84.1}} & \textbf{69.5} & \textbf{48.3}
\\ %
 \midrule %
{\color{teal}\scriptsize 12} & \note{DrostPPF~\cite{drost2010model}}& $\greencheckmark$ & $\greencheckmark$  & $\greencheckmark$ & $\greencheckmark$ & \note{52.7} & \note{-} & \note{-} & \note{-} & \note{-} & \note{-} & \note{34.4} & \note{-} & \note{-} \\ %

{\color{teal}\scriptsize 13} & \note{PPF + Zephyr~\cite{okorn2021zephyr}}& $\greencheckmark$ & $\greencheckmark$ & $\greencheckmark$ & $\greencheckmark$ & \note{59.8} & \note{-} & \note{-} & \note{-} & \note{-} & \note{-} & \note{51.6} & \note{-} & \note{-} \\ %

{\color{teal}\scriptsize 14} & OSOP~\cite{shugurov2022cvpr:osop}& $\greencheckmark$ & $\redxmark$  & $\greencheckmark$ & $\greencheckmark$ & 48.2 & - & - & - & - & 60.5 & 57.2 & - & - \\ %

 {\color{teal}\scriptsize 15} & Ours + MegaPose~\cite{labbe2022megapose}& $\greencheckmark$ & $\greencheckmark$  & $\greencheckmark$ & $\greencheckmark$ & \textbf{60.1} & 46.8 & \textbf{84.3} & 32.7 & 47.9 & \textbf{68.6} & 75.7 & 59.4 & 234.1 \\ %
 {\color{teal}\scriptsize 16} & \note{Ours (Pyrender)} & $\greencheckmark$ & $\greencheckmark$ & $\greencheckmark$ & -- & \note{49.0} & \note{39.6} & \note{74.7} & \note{30.3} & \note{37.3} &\note{56.1}& \note{66.6} & \note{50.5} & \note{4.95} \\ %
 {\color{teal}\scriptsize 17} & Ours & $\greencheckmark$ & $\greencheckmark$ & $\greencheckmark$ & -- & 50.9 & 42.1 & 60.4 & 42.0 & 46.3 & 53.8 & 64.8 & 51.5  & \textbf{4.81}\\ %
 
{\color{teal}\scriptsize 18} & Ours & $\greencheckmark$ & $\greencheckmark$ & $\greencheckmark$  & $\greencheckmark$ & 58.3 & \textbf{49.6} & 72.5 & \textbf{44.9} & \textbf{51.5} & 64.0 & \textbf{79.0} & \textbf{60.0}  & 85.9\\ %
\bottomrule
\end{tabular}
  
\end{adjustbox}
\label{tab:bop}
\end{table*}

\subsection{Benchmark}
\textbf{Datasets}. For the training dataset, we take the GSO dataset ~\cite{downs2022google} to pretrain the Geo model and the geometric feature matching module, which contains 1000 3D objects under household scenes and 1 million synthetic images provided by Megapose~\cite{labbe2022megapose}.
\note{For the test datasets, we follow the BOP challenge~\cite{hodavn2020bop} to select the seven core datasets where all test images are captured from the real-world environment, including LineMod Occlusion (LMO), T-LESS, TUD-L, IC-BIN, YCB-Video (YCB-V), ITODD, and HomebrewedDB (HB). The first 5 datasets are open for offline access and typically adopted as the ablation study benchmark, named BOP 5. The 7 core datasets are denoted BOP 7. 
BOP datasets include over a hundred objects with a variety of shapes and textures and twenty thousand images from challenging real scenes.}

\textbf{Metrics.}
For the instance segmentation metric, we adopt the Mean Average Precision (mAP) as defined in the BOP Challenge~\cite{hodavn2020bop, coco, bop2023}.
\note{For the pose estimation metric, we follow~\cite{hodavn2020bop, bop2023} to measure the Average Recall (AR) of the mean of VSD (Visible Surface Discrepancy), MSSD (Maximum Symmetry-Aware Surface Distance), and MSPD (Maximum Symmetry-Aware Projection Distance) to measure the pose estimation performance performance of pose estimation. More detailed definitions of evaluation metrics can refer to the BOP Challenge~\cite{hodavn2020bop, bop2023}.}

\subsection{Implementation Details}
\label{sec: impl}
ZeroPose is trained on the SLURM cluster with 8 NVIDIA V100 and inferring on the PC with NVIDIA RTX 3090.

\label{sec:implementation_details}
\textbf{Visual model.} We adopt the pretrained visual foundation models DINOv2~\cite{dinov2} as our visual model, which is based on the ViT architecture~\cite{vit}.  The visual features of foreground instances are cropped and resized with shape (224, 224, 3) as input to extract the scene images template images embeddings, and patch embeddings. For the proposed text-prompt zero-shot instance segmentation pipeline, we substitute the visual model with a multi-modal vision model~\cite{imagebind} and replace the template image embedding with text embedding derived from the given object name. For datasets without object names or text descriptions, we generate a caption from the 3D-text multi-modal model PointLLM~\cite{xu2023pointllm} as the text description.

\textbf{Geo model.} To improve the robustness of geometric extraction and matching at the proposed normalized point cloud pairs, 
we pretraining the Geo model and geometric matching model on the synthetic GSO~\cite{downs2022google, labbe2022megapose} dataset.
To enhance robustness against variations in illumination and occlusions, our approach incorporates data augmentation techniques, including color and position jittering.
For the training of high-level feature extraction and matching, we adopt the overlap-aware circle loss~\cite{qin2023geotransformer}, to train the model to locate the high overlap region.
We calculate the correspondence regions by the ground truth pose and select the overlap of correspondence regions more than 10\% as positive samples. The others that have an overlap of less than a threshold are viewed as negative samples.
For low-level, we match each low-level point pair by the distance within a matching radius and adopt negative log-likelihood loss to train the network to find the correct matched pairs. 
High-level and low-level features can be simultaneously trained in a network architecture with an equal weight based on their respective loss functions.

\textbf{Segmentation anything.} We adopt the recent lightweight version interactive segmentation method FastSAM~\cite{fastsam}, with default CNN backbone to generate the mask of foreground instances and filter out the tiny regions with less than 128 pixels. 
Moreover, since there are possible multiple point prompts for one instance, we remove the duplicating regions in predicted foreground instances by the Non-Maximum Suppression algorithm to filter the Intersection over Union (IOU) of masks over 0.25 for each object.

\subsection{Evaluation of Zero-Shot Instance Segmentation}
\label{sec:exp_seg}
Since there are no other related methods following the CAD-model-prompt zero-shot instance segmentation setting, we select two classic supervised object-specific instance segmentation methods and implement a text-prompt zero-shot instance segmentation method for comparison. 
Compared with the object-specific methods, although without the training for target objects, the proposed object discovery module achieves comparable performance with the classic Mask RCNN method~\cite{maskrcnn}. However, compared with the state-of-the-art object-specific method ZebraPoseSAT~\cite{su2022zebrapose}, there is still room for improvement.
Moreover, for a fair comparison of different prompts, we implement a text-prompt zero-shot instance segmentation pipeline following the same pipeline as in our discovery step. We substitute the visual model with a multi-modal vision model~\cite{imagebind} and replace the template image embedding with text embedding derived from the given object name.
As depicted in Table~\ref{tab:ZSIS result}, our CAD-model-prompt zero-shot instance segmentation method shows obvious performance improvements compared with the text-prompt method from 15.0\% to 30.4\%. When converting the multi-modal vision model Imagebind~\cite{imagebind} into the visual foundation modal DINOv2~\cite{dinov2}, the performance is increased to 37.8\%. These demonstrate the proposed CAD model prompt is more reliable than the text prompt in the object pose estimation scenario.

\subsection{Evaluation of Zero-Shot Pose Estimation}
To evaluate our pose estimation performance, we compared classic object-specific pose estimation methods in Table~\ref{tab:bop} rows 1-4 and \note{the latest partial (rows 5-11) or fully (rows 12-17) zero-shot pose estimation methods, where the partial zero-shot methods adopt an object-specific object discovery model training on the prompted objects.}
\note{
Compared with classic object-specific methods, the proposed ZeroPose without model retraining on specific test objects outperforms the CosyPose~\cite{cosypose} and CDPNv2~\cite{li2019cdpn}.
For the ADD-S 2CM metric in the DenseFusion~\cite{densefusion}, under the same segmentation results, our method achieves 96.6 in the YCB-V~\cite{posecnn} dataset, which is comparable with the 96.8 in DenseFusion~\cite{densefusion} training on the specific test objects.}
However, as a zero-shot method, it still has room for improvement compared with the recent object-specific methods. 

\note{Compared with zero-shot methods, ZeroPose demonstrates better accuracy and generalization capability under both the partial and fully zero-shot settings. The Discovery-Orientation-Registration (DOR) pipeline enables to perform on a variety and challenging scenes such as cluttered scenes and generalize to unseen objects.
OVE6D~\cite{cai2022ove6d}, GCPose~\cite{gcpose}, and Megapose~\cite{labbe2022megapose} are capable of performing unseen object pose estimation without retraining. However, these methods require human intervention to segment the target objects from the scene image, preventing them from operating in a fully zero-shot manner. 
These methods evaluate the performance of the BOP datasets requiring a customized Mask-RCNN model~\cite{maskrcnn} to segment the objects of interest and then perform pose estimation with these accurate masks.
For fairly compared with them, we adopt the same object-specific training Mask RCNN model~\cite{maskrcnn} for object discovery and compare the pose estimation performance.}
As demonstrated in rows 5 and 7, ZeroPose outperforms the state-of-the-art method MegaPose and achieves 50 times running speed improvement and 98\% running time reduction.
Since the existing pose refinement strategies~\cite{lipson2022coupled, svd, labbe2022megapose} are typically generalized to novel objects, pose initialization is the biggest challenge in zero-shot object pose estimation.
Compared to MegaPose~\cite{labbe2022megapose} for pose initialization under the same instance segmentation results provided by the MaskRCNN~\cite{maskrcnn}, our method achieves a 44.9\% performance gain and is 14.2 times faster, as shown in rows 5 and 7 of Table~\ref{tab:bop}.
Moreover, the MegaPose has a limitation in that it requires a detection result from the user as input (supervised Mask RCNN provided in the BOP evaluation) to locate the candidate object but our method can estimate the pose in a fully zero-shot manner with comparable performance of 51.5\%. 

In addition, we also integrate our zero-shot object discovery into MegaPose. As shown in row 15, the MegaPose equipped with our object discovery surpasses its original version (59.4\% vs 57.2\%), demonstrating the effectiveness of our object discovery. To evaluate the optimal performance of the zero-shot method, we integrated the refiner from MegaPose~\cite{labbe2022megapose} with the proposed ZeroPose framework. As illustrated in row 18, it achieves 60\% for AR outperforming the object-specific methods, demonstrating the effectiveness of the zero-shot method.

\note{Compared to the non-learning-based method DrostPPF~\cite{drost2010model} and its refinement method Zephyr~\cite{okorn2021zephyr}, our ZeroPose achieves better average results and robustness, which can perform at more challenging datasets.}
Compared to OSOP~\cite{shugurov2022cvpr:osop}, although its object discovery module is zero-shot, it introduces a presupposing scene condition of one instance per object, limiting its applicability to the cluttered scene with multiple instances of the same object. 
Even without the presupposing scene condition, ZeroPose surpasses OSOP on the mean of their provided three datasets (55.3\% and 56.5\%) and shows comprehensive performance surpassing after adopting the refiner~\cite{labbe2022megapose}. As shown in rows 9 and 11, for datasets (T-LESS, IC-BIN, ITODD) with cluttered scenes, OSOP is unavailable, but the proposed ZeroPose achieves both zero-shot object discovery and pose estimation. These experiment results and comparisons demonstrate the effectiveness and efficiency of the proposed zero-shot object pose estimation method.

\subsection{Visualization}
\begin{figure*}[t]
    \centering
    \includegraphics[width=0.99\textwidth]{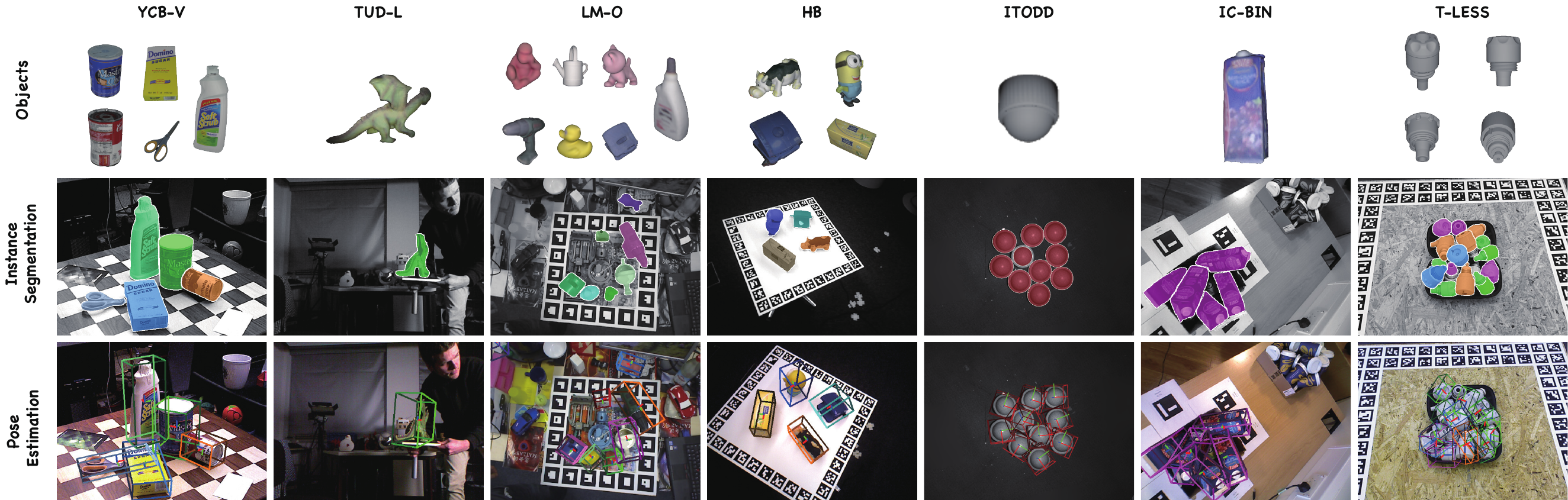}
    \caption{\note{Visualization instance segmentation and pose estimation results on the 7 BOP core datasets. The first row represents the prompted objects.}}
    \label{fig:vis}
\end{figure*}
As illustrated in Figure~\ref{fig:vis}, ZeroPose performs well on a variety of objects, from simple geometric shapes (as seen in LM-O and TUD-L) to more complex and textured objects (as seen in YCB-V). Furthermore, as demonstrated by the object discovery (instance segmentation) and pose estimation results in the ITODD, IC-BIN, and ITODD datasets, ZeroPose exhibits reasonably good performance in cluttered scenes. For objects with less distinctive features or similar shapes (YCB-V and T-LESS datasets), ZeroPose can distinguish objects based on texture information and local geometric structure differences. 
The visualization analysis indicates that ZeroPose can handle a diverse set of novel objects and scene conditions.

\begin{figure}[t]
    \centering
    \includegraphics[width=0.49\textwidth]{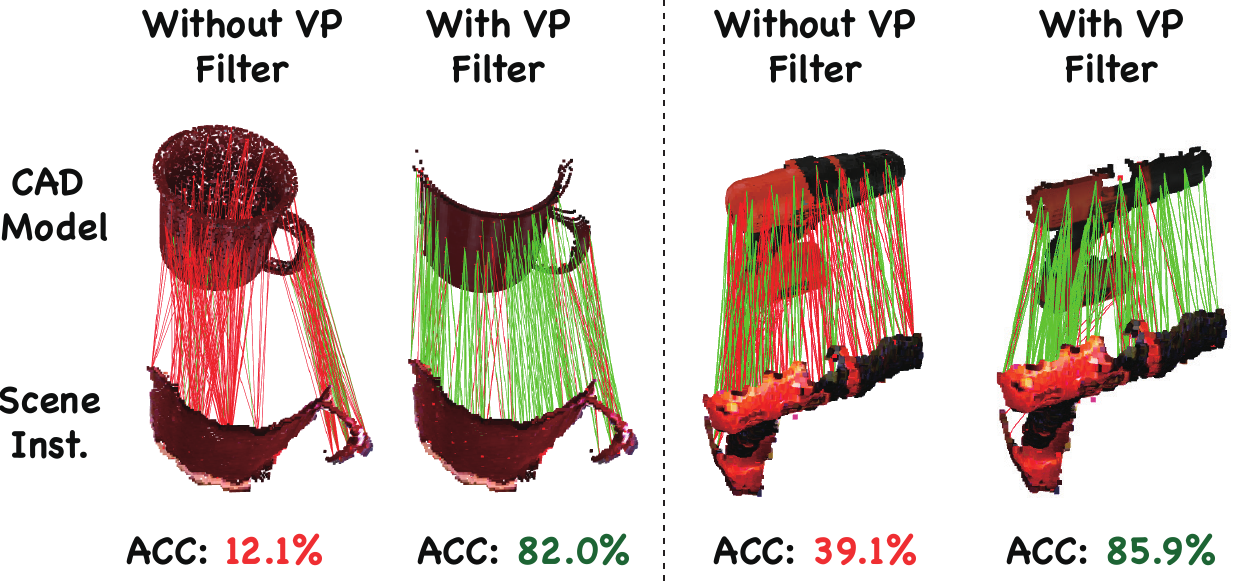}
    \caption{Visualization of the geometric matching. The first and third columns are complete CAD model point clouds and the second and fourth columns are viewpoint-filtered point clouds. ACC stands for feature matching accuracy.}
    \label{fig:vis_vp}
\end{figure}
\subsection{Ablation Study}
\begin{table*}[t]
  \centering
  % \small %
 \setlength{\tabcolsep}{3.5pt}
\caption{\note{Comparison of the viewpoint densities}. 
The metric is mAP~\cite{hodavn2020bop} for instance segmentation task.}
\label{tab:aba-renderviews}
\begin{adjustbox}{max width=\textwidth}

  \begin{tabular}{ccccccccccccc}
 \toprule  
& \multicolumn{3}{c}{Onboarding Time (s)} & \multicolumn{1}{c}{Memory (MB)} & \multicolumn{5}{c}{BOP5 Datasets} & & \\ 
 \cmidrule(lr){2-4} \cmidrule(lr){5-5} \cmidrule(lr){6-10}
   Template Num. & {\scriptsize \makecell[b]{Render}} & {\scriptsize \makecell[b]{Feat Extraction}} & {\scriptsize \makecell[b]{Total}} & {\scriptsize \makecell[b]{Template Feat.}} & \textsc{lm-o} & \textsc{t-less} & \textsc{tud-l} & \textsc{ic-bin} & \textsc{ycb-v} & Mean & Time \note{(s)} \\

 \midrule %
  6 & 0.3 & 0.1 & 0.4 & 0.02 & 29.7 & 11.6 & 45.2 & 19.8 & 32.0 & 27.6 & 0.240  \\ %
   42 & 26.5 & 0.5 & 27.0 & 0.16 & 37.7 & 34.7 & 46.0 & 20.6 & 57.4 & 39.3 & 0.242  \\ %
  162 & 102.2 & 2.3 & 104.5 & 0.64 & 37.8 & 35.7 & 45.4 & 20.8 & 56.0 & 39.2 & 0.251  \\ %
   642 & 405.1 & 7.2 & 412.3 & 2.51 & 37.4 & 36.4 & 44.7 & 20.5 & 56.0 & 38.9 & 0.289  \\ %
 
  \bottomrule
  \end{tabular}
  
\end{adjustbox}
\vspace{+10 pt}
\end{table*}

\begin{table}[t]
    \centering
    % \small %
   \setlength{\tabcolsep}{3.5pt}
  \caption{Comparison of viewpoint prediction strategies. The metric is the degree for the predicted viewpoint direction with the Ground Truth direction. The degree error is from 0 to 180, where the lower value indicates a higher accuracy in predicting the viewpoint direction. \note{D step indicates the result from the first discovery step.}}
  \label{tab:aba-vp}
  \begin{adjustbox}{max width=\textwidth}
  
    \begin{tabular}{lccccccc}
   \toprule  
     Setting & \textsc{lm-o} & \textsc{t-less} & \textsc{tud-l} & \textsc{ic-bin} & \textsc{ycb-v} & Mean  \\
      \midrule
    Image-Level \note{(D step)}& 57.3 & 45.6 & 51.7& 74.9& 55.9 & 57.1\\
    Patch-Level (top1) & 41.2 & 34.7 & 34.3 & 63.6 & 36.1 & 42.0 \\
    Patch-Level (top5) & \textbf{19.5}& \textbf{21.5}& \textbf{15.5}& \textbf{26.4}& \textbf{18.9}& \textbf{20.4}\\   
    \bottomrule
    \end{tabular}
    
  \end{adjustbox}
  \end{table}

\begin{table}[t]
\centering
\caption{\note{Comparison of different pose estimation strategies. HGM is the Hierarchical Geometric Matching model. IP Rot. denotes the template viewpoints with additional in-plane rotations}}

\begin{adjustbox}{max width=\textwidth}

\begin{tabular}{cccc}
\toprule
     & Model & Paradigm  &  AR (\%)\\\midrule
{\color{teal}\scriptsize 1} & Visual & Feature Matching & 16.3 \\
{\color{teal}\scriptsize 2} & Visual & Template Matching & 24.4 \\
{\color{teal}\scriptsize 3} & Visual & Template Matching + ICP~\cite{ICP}  & 22.9 \\
{\color{teal}\scriptsize 4} & Visual & Template Matching + IP Rot.  & 29.9 \\
{\color{teal}\scriptsize 5} & Geo & Feature Matching & 49.1 \\
{\color{teal}\scriptsize 6} & Geo + HGM & Feature Matching & 70.4\\
\bottomrule  
\end{tabular}
\end{adjustbox}
\label{tab:aba-pose-strategies}
\end{table}

\begin{table}[t]
  \centering
  % \small %
 \setlength{\tabcolsep}{3.5pt}
\caption{{Comparison of pose estimation at different viewpoint-filtered CAD model points. W/O VP is the CAD model without filtering from the predicted viewpoint. TOP N represents adopting the highest N score of predicted viewpoints to filter point clouds.}}
\label{tab:aba-vp-pose}
\begin{adjustbox}{max width=\textwidth}

  \begin{tabular}{lccccccc}
 \toprule  
   Setting & \textsc{lm-o} & \textsc{t-less} & \textsc{tud-l} & \textsc{ic-bin} & \textsc{ycb-v} & Mean  &  Time (s)\\
  \midrule
  W/O VP & 50.2 & 37.9 & 86.8& 51.0& 65.3 & 58.2 & 0.058 \\
  TOP1 & 53.0 & 44.3 & 81.4 & 44.0 & 63.1 & 57.2 & 0.053 \\
  TOP2 & 61.7 & 55.1 & 89.3 & 54.2 & 72.3 & 66.5 & 0.080 \\
  TOP3 & 64.6 & 59.3 & 90.8 & 56.3 & 73.9 & 69.0 & 0.108 \\
  TOP4 & 66.1 & 61.5 & 91.9 & 56.9 & 73.4 & 70.0 & 0.138 \\
  TOP5 & 67.1 & 62.6 & 92.7 & 56.8 & 73.0 & 70.4 & 0.165 \\
  TOP6 & 67.5 & 63.7 & 93.2 & 56.6 & 72.8 & 70.8 & 0.191 \\
  \bottomrule
  \end{tabular}
  
\end{adjustbox}
\end{table}

\textbf{Comparison of the viewpoint densities.}
To evaluate the effect of different viewpoint densities, we render the template images from different viewpoint densities and viewpoint sampling strategies. For six template images, the camera positions encompass the front, back, left, right, top, and bottom orientations relative to the object.
The others are uniformly sampled from the surface of a unit sphere with different densities following~\cite{nguyen2022template}.
As shown in Table~\ref{tab:aba-renderviews}, we compare the onboarding time, cache memory consumption, inference running time, and accuracy (mAP) at different viewpoint densities.
Since the 6 template images are too sparse, they are unable to fully cover possible object appearances. 
Additionally, experimental results indicate that excessively dense template images do not yield additional performance gains in our setting. This is because the viewpoint angular difference between the template images and the scene images is already minimal, which is no more than 15.9 degrees in the 42-viewpoints setting. Consequently, the use of excessively dense template images not only fails to enhance the discriminability of positive samples but also increases the confidence scores of confounding negative samples within the scene, thereby impeding object discovery.
Hence, we choose 42 viewpoints template images for a CAD model in this paper.

\note{\textbf{Comparison of the template images rendering engines.}
To evaluate the effect on the template images render engine, we compare a high-fidelity engine Blender~\cite{denninger2019blenderproc}, and a high-speed engine, Pyrender~\cite{Pyrender}.
The rendered template images from Blender have better illumination consistency with the scene images compared with the high-speed engine.
Compared with Blender, Pyrender only leads to marginal performance drops of 1\% in average AR for pose estimation in line 8 of Table~\ref{tab:bop} and 0.7\% mAP in line 16 of Table~\ref{tab:ZSIS result} for instance segmentation. Namely, Pyrender saves 95\% of preprocessing time without significantly compromising pose estimation performance. For large-scale applications, we can use Pyrender as the rendering engine. In this way, the proposed method not only performs online pose estimation efficiently but also conducts offline preprocessing at a low time cost.}

\begin{figure}[t]
	\centering
 	\begin{minipage}{0.49\linewidth}
		\centering
		\includegraphics[width=0.99\linewidth]{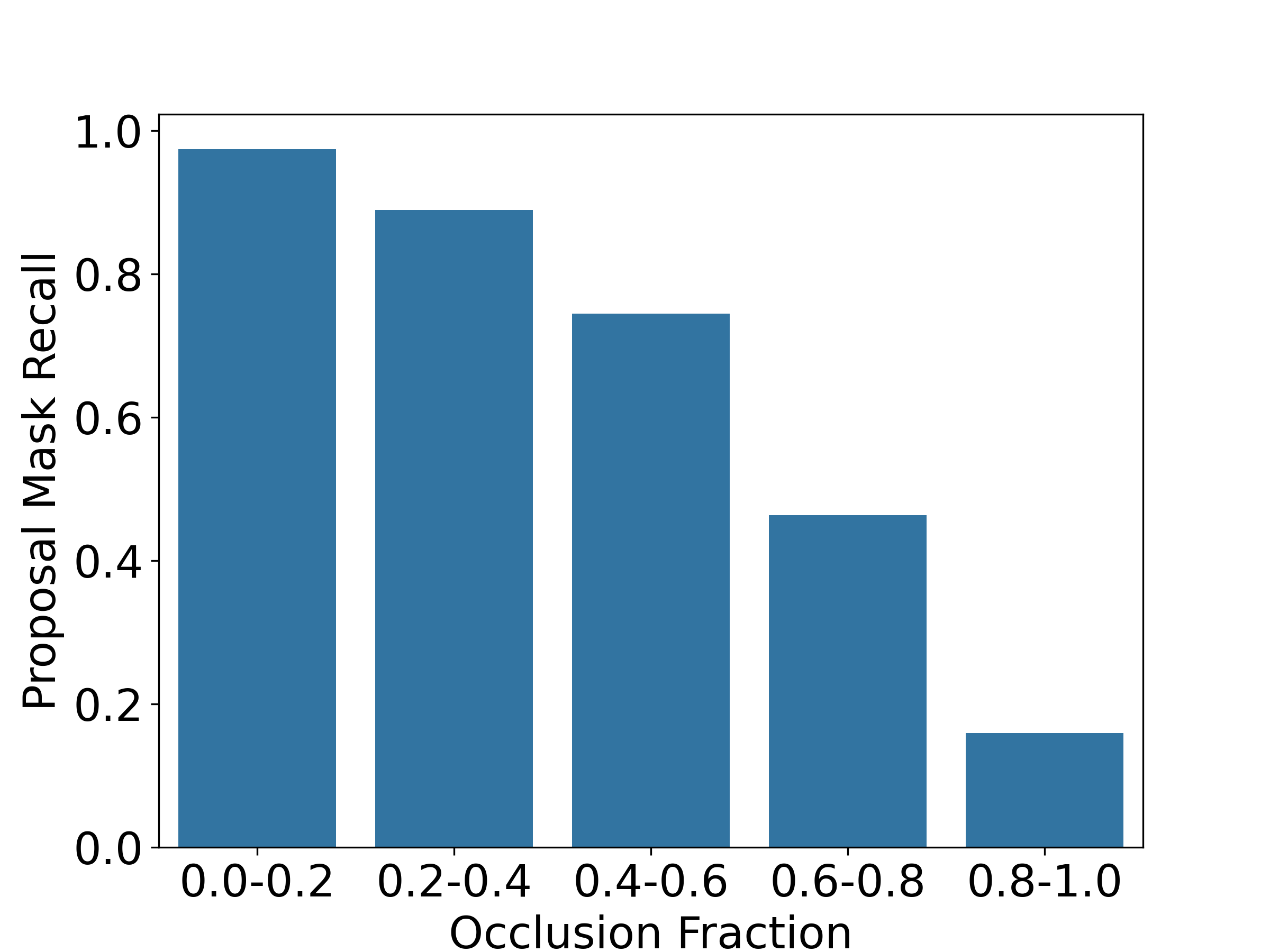}
		\caption{\note{Proposal object-agnostic instance segmentation masks recall at $IOU>0.5$.}}
		\label{fig:r2_box_recall}
	\end{minipage}
	\begin{minipage}{0.49\linewidth}
		\centering
		\includegraphics[width=0.99\linewidth]{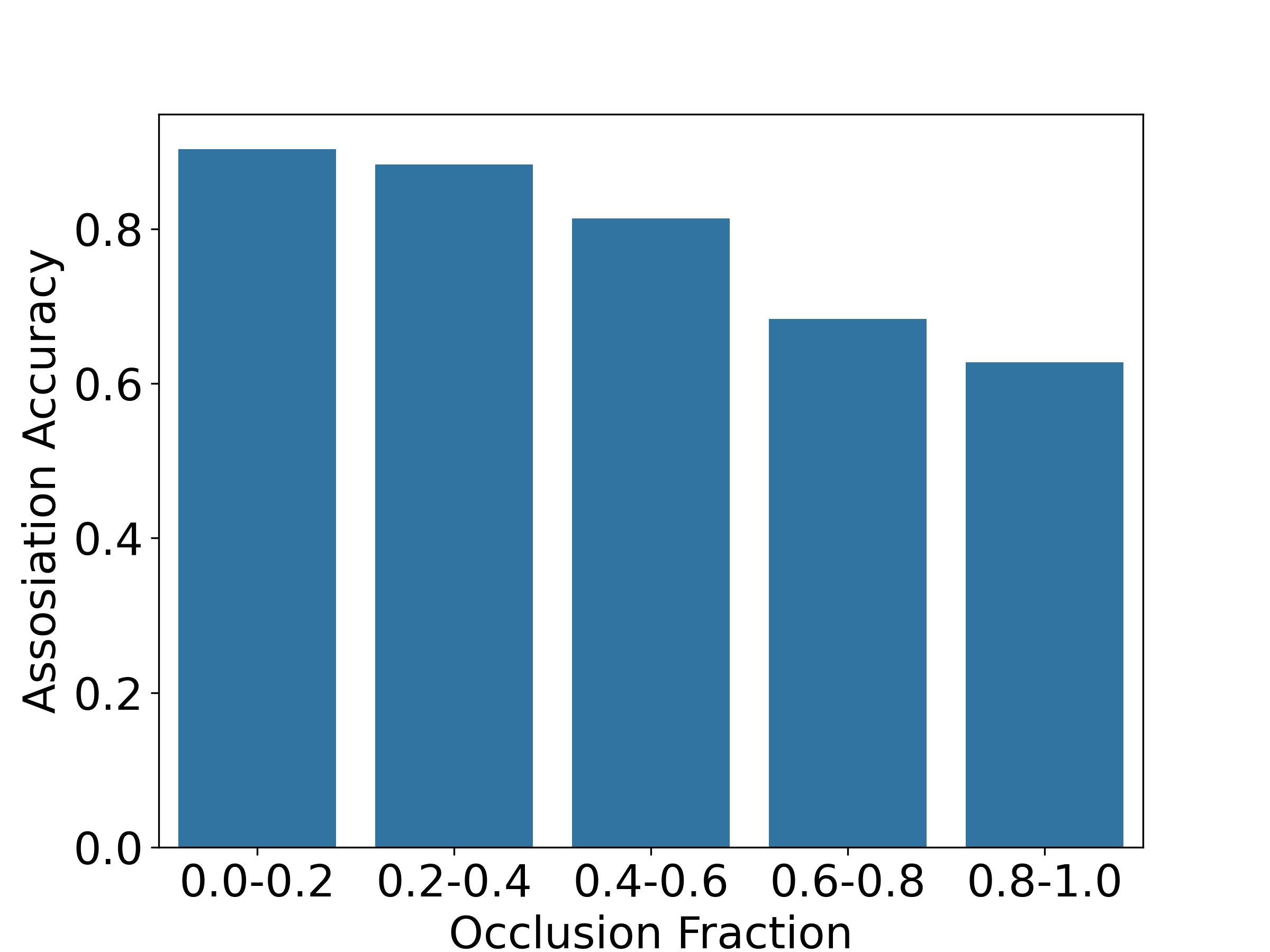}
		\caption{\note{Accuracy of association proposal masks and templates at different instance visibility fractions.}}
		\label{fig:r2_asso_acc}
	\end{minipage}
\end{figure}

\begin{table}[t]
  \centering
  % \small %
 \setlength{\tabcolsep}{6.5pt}
\caption{\note{Instance segmentation mAP using the score from different steps over five BOP datasets. }}
\label{tab:aba-score-revise}
\begin{adjustbox}{max width=\textwidth}

  \begin{tabular}{lcccccc}
 \toprule  
   Step &{\textsc{lm-o}} & {\textsc{t-less}} & \multicolumn{1}{c}{\textsc{tud-l}} & \multicolumn{1}{c}{\textsc{ic-bin}} & \multicolumn{1}{c}{\textsc{ycb-v}} & Mean \\

 \midrule %
 Discovery  & 37.7 & 34.7 & 46.0 & 20.6  & 57.4 & 37.8\\ %
 Orientation  & 40.0 & 37.1 & 47.2 & 22.8  & 57.4 & 40.9 \\
  \bottomrule
  \end{tabular}
  
\end{adjustbox}
\end{table}

\note{\textbf{Effect of occluded object discovery.} To analyze the robustness of the object discovery under occlusion, we calculate the segmentation recall of generated proposals having an IoU over 0.5 with Ground Truth and the association accuracy of proposals on 5 core datasets of BOP (BOP5) under different occlusion fractions, as shown in Figure~\ref{fig:r2_box_recall}. When the object is half occluded (0.4-0.6 occlusion fraction), the segmentation recall is over 70\% and the association accuracy is over 80\%. Heavy occlusion (\textit{e.g.}, 0.8-1.0 occlusion fraction) will affect the segmentation recall and association accuracy, as too few pixels are visible to recognize the target object.}

\note{\textbf{Effect of over/under-segmentation filtering.}
We also conducted experiments to analyze the effectiveness of the scoring and filtering strategies to filter over/under-segmentation. We found that 88\% of the under-segmentation masks and 74\% of the over-segmentation masks can be correctly filtered out. }

\textbf{Comparison of viewpoint prediction strategies.}
To evaluate the effect of viewpoint prediction strategies, we utilize the metric of cosine degree (range from 0 to 180) between the predicted viewpoint direction and the ground truth viewpoint direction.
Compared with the image embedding matching in the discovery stage, the proposed viewpoint prediction from the patch embedding matching strategy reduces the degree of viewpoint direction error from 57.1 to 42 degrees, as shown in Table~\ref{tab:aba-vp}. 
Furthermore, when selecting the top 5 candidate viewpoints, the error decreases to 20.4 degrees, approaching the minimal angle error of viewpoint direction in the templates (15.9 degrees). This demonstrates the effectiveness of the proposed camera observation viewpoint prediction.

\note{Considering that the global feature may be sensitive to occlusions, we leverage the patch-level feature to revise the score of predicted instances. As shown in Table~\ref{tab:aba-score-revise}, the score revision method leads to a mean performance gain of 3.1\% in mAP over 5 BOP datasets, demonstrating its effectiveness.}

\textbf{Comparison of pose estimation strategies.}
To compare the effect of the different pose estimation strategies, we implement four distinct pose estimation strategies. 
The first strategy utilizes a visual model to extract patch embeddings for feature matching, subsequently estimating the pose through the RANSAC-Perspective-n-Points algorithm~\cite{pnp}. As delineated in Section~\ref{sec:method_pose}, this strategy is unreliable in the object pose estimation scenario, showing a mere 16.3\% average AR across the BOP 5 datasets. 
\note{The second strategy adopts the template pose from the Orientation step as rotation and estimates the translation component, as described in \cite{nguyen2022template}, for the final pose estimation.
Based on the second strategy, the third and fourth strategies further introduce an ICP algorithm and templates across different in-plane rotations (ten samples with a density of 36 degrees), thereby generating a candidate rotation for each template. Since the difference between the predefined 2 DoF viewpoint and 6 DoF camera perspective, the viewpoint prediction is unable to perniciously predict the object pose.
}

When instead of the visual model in the proposed Geo model, the feature matching by geometric embedding results in an obvious improvement in performance relative to the visual patch embedding feature matching paradigm. Furthermore, the integration of a hierarchical geometric matching model enhances the reliability of the matching process, leading to an additional performance gain.

\textbf{Effect of viewpoint-filtered CAD model point cloud and normalized representation.}
We compare the CAD model point cloud with or without filtering the self-occlusion region by viewpoint prediction.
As illustrated in Figure~\ref{fig:vis_vp}, there are many ambiguous regions in the CAD model leading to mismatching between the points in the scene instances and the ambiguous self-occlusion regions in the CAD model. The estimated camera viewpoint effectively finds the points of the CAD model corresponding to the visible points of the scene instance and filters out the remaining points of the CAD model, improving the matching accuracy.

Moreover,  we quantitatively analyze the impact of the viewpoint-filtered CAD model point cloud on pose estimation in Table~\ref{tab:aba-vp-pose}. The AR performance markedly increases when adopting more than two candidate viewpoint-filtered point clouds. For the number of viewpoint candidates, although fewer viewpoint-filtered point clouds result in faster running time, it is sensitive to the viewpoint prediction result. More candidate point clouds can improve the robustness of inaccuracy viewpoint prediction and advanced performance gain. To balance speed and accuracy, we choose the top 5 candidates for viewpoint-filtered point cloud in this paper.

To validate the effectiveness of normalized representation point clouds, we compared scene and object point clouds with or without the proposed normalization.
As demonstrated in Table~\ref{tab:scale}, the normalized representation point clouds can enhance the performance whether under the completed point clouds (+ 9.9\%) or viewpoint-filtered point clouds (+ 5.1\%), verifying the effectiveness of normalized representation.

\begin{table}[t]
\centering
      \caption{Comparison of point cloud input and representation. Viewpoint Filtered represents the CAD model point cloud filtered by predicted camera observation viewpoint. Normalized Rep. represents the point cloud that is scaled into the normalized representation. AR is the average on the BOP5 datasets.}
      \begin{adjustbox}{max width=\textwidth}

    \begin{tabular}{cccc}
      \toprule
         & Viewpoint Filtered & Normalized Rep. & AR (\%) \\
        \midrule
        {\color{teal}\scriptsize 1} & - & - & 48.3\\
        {\color{teal}\scriptsize 2} & - & $\greencheckmark$ & 58.2\\
        {\color{teal}\scriptsize 3} & $\greencheckmark$ & - & 65.1\\
        {\color{teal}\scriptsize 4} & $\greencheckmark$ & $\greencheckmark$ & 70.4\\
      \bottomrule  
  \end{tabular}
    \end{adjustbox}
    \label{tab:scale}
\end{table}

\section{Conclusion}
\label{sec:conclusion}
In this paper, we propose a universal framework ZeroPose, which solves both object discovery and pose estimation
in a zero-shot manner without additional human interaction or presupposing scene conditions. It performs pose estimation in a novel Discovery-Orientation-Registration (DOR) inference pipeline through stepwise feature matching across three distinct inference steps. 	
For limitation, the three steps in the DOR pipeline are independent and can not be trained end-to-end. An end-to-end pipeline can learn the mapping from input data to pose estimation output directly, simplifying the inference pipeline. However, since the end-to-end pipeline is fully data-driven, there is a lack of large volumes of labeled real data for training. 
For further work, a large-scale real-world pose estimation dataset with large volumes of labeled data is valuable for improving the performance and generalization of the model.
Additionally, exploring multi-task learning frameworks that integrate related object discovery tasks including object detection, tracking, instance segmentation, and pose estimation is a promising research topic.

\bibliographystyle{IEEEtran}
\bibliography{ref.bib}

\begin{IEEEbiography}[{\includegraphics[width=1in,height=1.25in,clip,keepaspectratio]{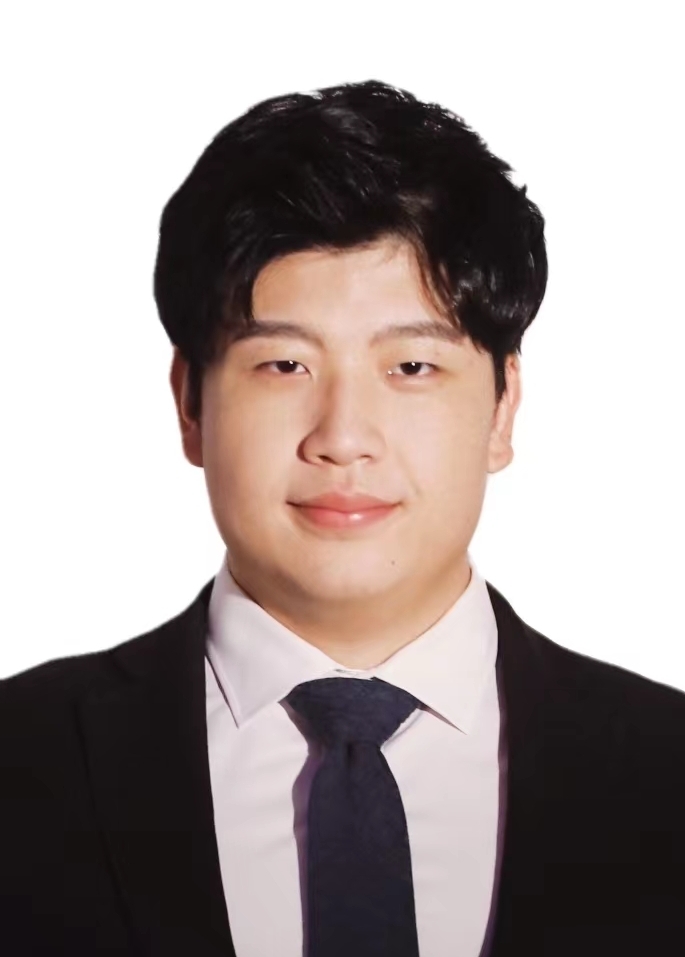}}] {Jianqiu Chen} received the M.S. degree from University of New South Wales in 2021. He is currently pursuing the Ph.D. degree with the School of Computer Science and Technology, Harbin Institute of Technology (Shenzhen), Shenzhen. His research interests include object pose estimation and zero-shot learning.
\end{IEEEbiography}

\begin{IEEEbiography}[{\includegraphics[width=1in,height=1.25in,clip,keepaspectratio]{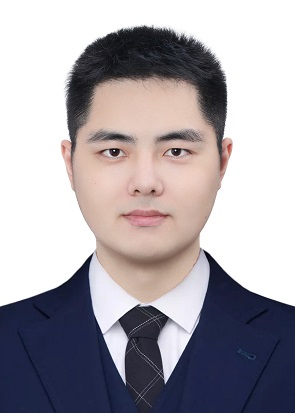}}] {Zikun Zhou}
received his Ph.D. and Master's degree from Harbin Institute of Technology in 2022 and 2018, respectively. He is currently an assistant research fellow with Pengcheng Laboratory. His research interests include computer vision and machine learning.
\end{IEEEbiography}

\begin{IEEEbiography}[{\includegraphics[width=1in,height=1.25in,clip,keepaspectratio]{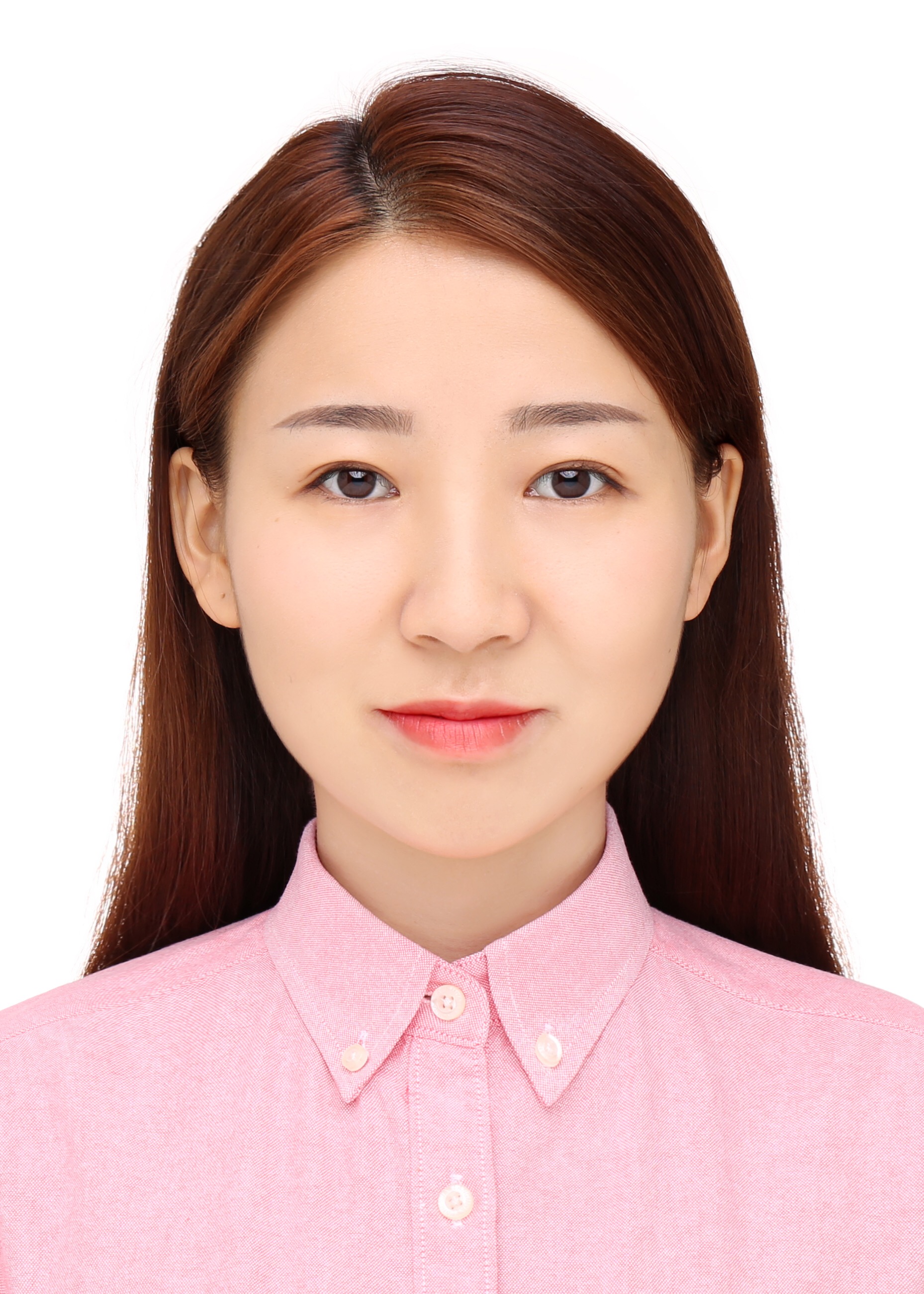}}] {Mingshan Sun}
 received the B.S. degree in Electronic Commerce from Guangxi University, Nanning, China, in 2017, and the M.S. degree in computer technology from the Harbin Institute of Technology, Shenzhen, China, in 2020. She is currently working at SenseTime Research. Her research interests include computer vision and machine learning.
\end{IEEEbiography}

\begin{IEEEbiography}[{\includegraphics[width=1in,height=1.25in,clip,keepaspectratio]{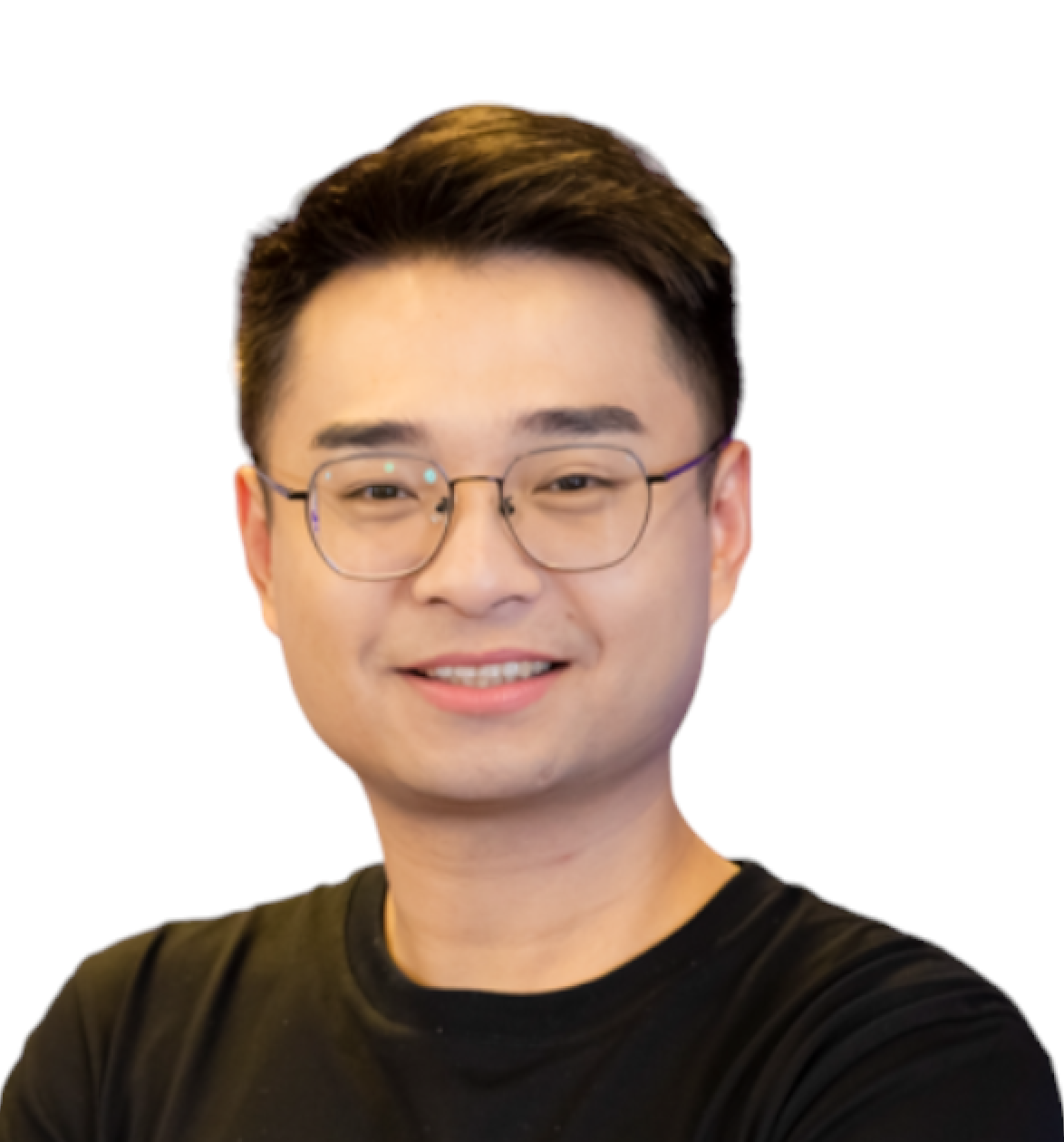}}]{Rui Zhao} received the B.S. degree from the University of Science and Technology of China in 2010 and the Ph.D. degree in electronic engineering from The Chinese University of Hong Kong in 2015. He joined a startup venture called SenseNets as the CTO right after postgraduate study. In 2018, he joined SenseTime Research, Shenzhen, Guangdong, China, as a Research Director, and has been the Head of the Smart City Group, Research and Development Department, SenseTime, since 2021. While in SenseTime Research, he led a team designing and developing competitive deep learning models and techniques, which are applied to products for smart city applications including public services, transportation, epidemic control, and smart manufacturing. He is currently an Adjunct Researcher at the Shenzhen Institute of Advanced Technology, Chinese Academy of Sciences, the Tsinghua Shenzhen International Graduate School, and the Qing Yuan Research Institute, Shanghai Jiao Tong University. His research interests span a range of topics in computer vision and deep learning, including face recognition, person re-identification, large-scale clustering, unsupervised/self-supervised learning, few-shot/zero-shot learning, and visual-language foundation models. He has published more than 60 technical papers and book chapters on these topics.

\end{IEEEbiography}

\begin{IEEEbiography}[{\includegraphics[width=1in,height=1.25in,clip,keepaspectratio]{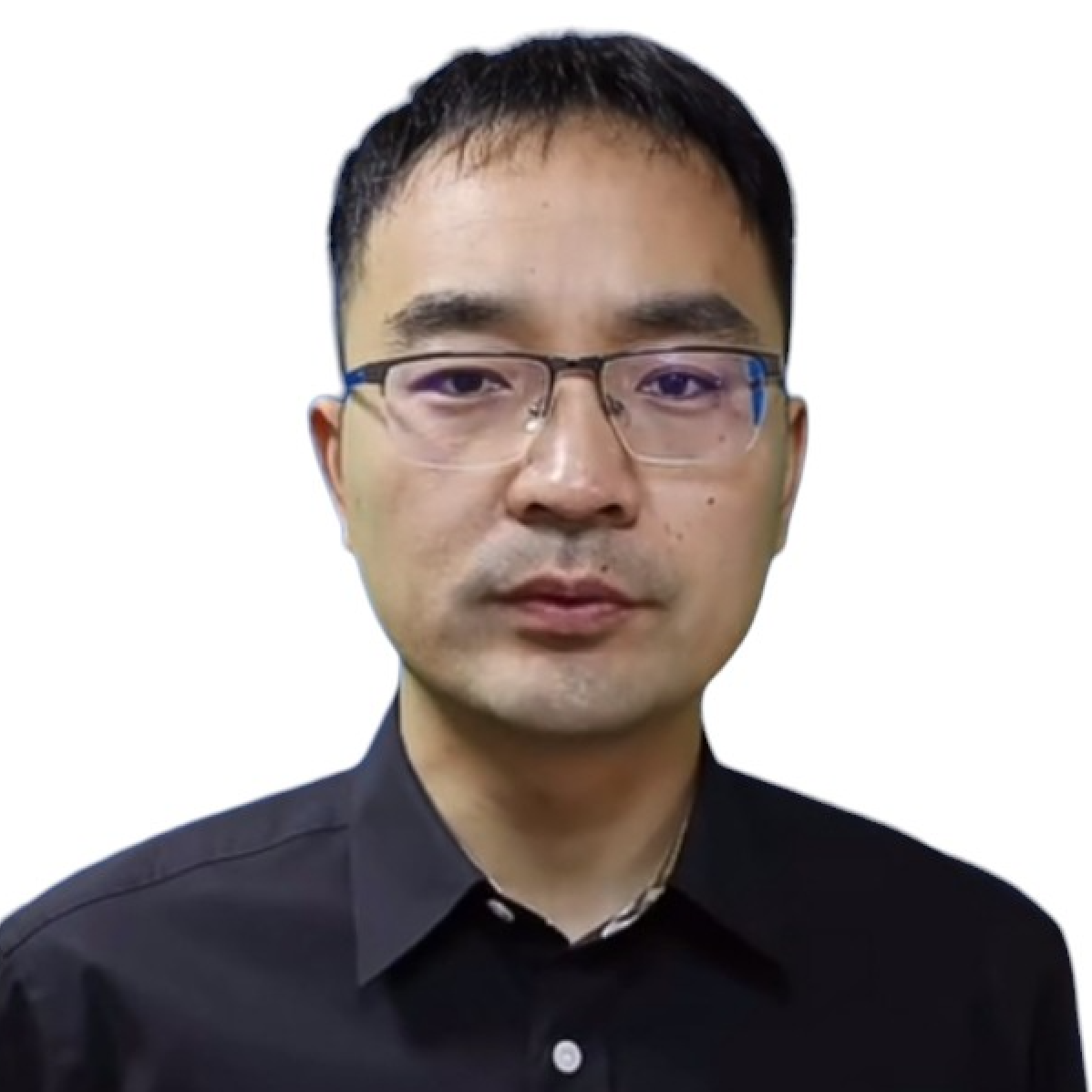}}] {Liwei Wu}
 received the B.S. degree in Automation from Nanjing University, Nanjing, China, in 2013 and the M.S. degree in Automation from Tsinghua University, Beijing, China, in 2016. He is currently working as a researcher in SenseTime Research. His research interests include computer vision and machine learning.
\end{IEEEbiography}

\begin{IEEEbiography}[{\includegraphics[width=1in,height=1.25in,clip,keepaspectratio]{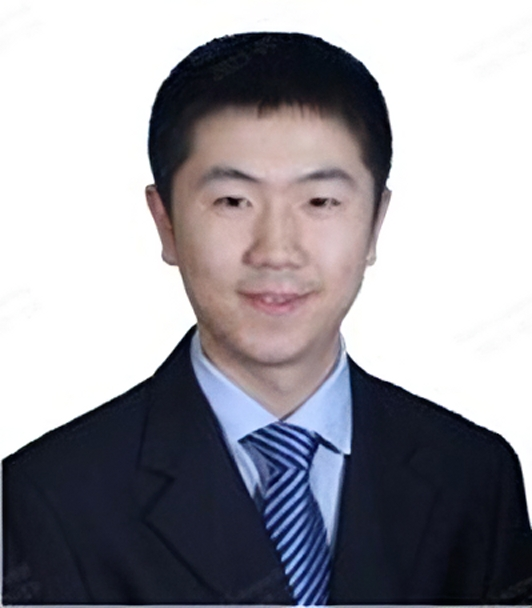}}] {Tianpeng Bao}
 received the B.S. degree in Automation from Tsinghua University, Beijing, China, in 2010 and the M.S. degree in Control Science and Engineering from Tsinghua University, Beijing, China, in 2014. He is currently working as a researcher in SenseTime Research. His research interests include computer vision and LLM-based agents.
\end{IEEEbiography}

\begin{IEEEbiography}[{\includegraphics[width=1in,height=1.25in,clip,keepaspectratio]{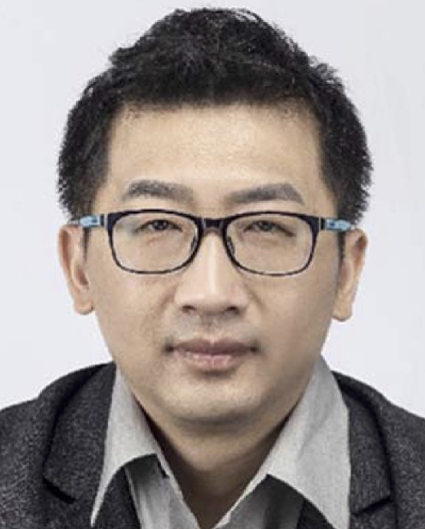}}] {Zhenyu He}
received his Ph.D. degree from the Department of Computer Science, Hong Kong Baptist University, Hong Kong, in 2007. From 2007 to 2009, he worked as a postdoctoral researcher in the department of Computer Science and Engineering, Hong Kong University of Science and Technology. He is currently a full professor in the School of Computer Science and Technology, Harbin Institute of Technology, Shenzhen, China. His research interests include machine learning, computer vision, image processing and pattern recognition.
\end{IEEEbiography}
\end{document}